\documentclass[preprints,article,accept,moreauthors,pdftex]{Definitions/mdpi} 
\firstpage{1} 
\pubvolume{1}
\issuenum{1}
\articlenumber{0}
\pubyear{2022}
\copyrightyear{2022}
\datereceived{} 
\dateaccepted{} 
\datepublished{} 
\hreflink{https://doi.org/} 

\usepackage{bm}
\usepackage[capitalise]{cleveref}
\usepackage{siunitx}
\usepackage{subcaption}
\usepackage{adjustbox}
\usepackage{pgfbaselayers}
\pgfdeclarelayer{edgelayer}
\pgfdeclarelayer{nodelayer}
\pgfsetlayers{edgelayer,nodelayer,main}
\usetikzlibrary{decorations,calligraphy}
\usepackage[flushleft]{threeparttable}

\Title{A machine learning and feature engineering approach for the prediction of the uncontrolled re-entry of space objects}

\TitleCitation{A machine learning and feature engineering approach for the prediction of the uncontrolled re-entry of space objects}

\Author{Francesco Salmaso $^{1}$, Mirko Trisolini \orcidA{} $^{1}$* and Camilla Colombo \orcidB{} $^{1}$}

\AuthorNames{Francesco Salmaso, Mirko Trisolini and Camilla Colombo}

\AuthorCitation{Salmaso, F.; Trisolini, M.; Colombo, C.}

\address[1]{%
$^{1}$ \quad Department of Space Science and Technology, Politecnico di Milano, via La Masa 34, 10156, Milan, Italy}

\corres{Correspondence: mirko.trisolini@polimi.it}

\abstract{
The continuously growing number of objects orbiting around the Earth is expected to be accompanied by an increasing frequency of objects re-entering the Earth’s atmosphere. Many of these re-entries will be uncontrolled, making their prediction challenging and subject to several uncertainties. Traditionally, re-entry predictions are based on the propagation of the object’s dynamics using state-of-the-art modelling techniques for the forces acting on the object. However, modelling errors, particularly related to the prediction of atmospheric drag may result in poor prediction accuracies.
In this context, we explore the possibility to perform a paradigm shift, from a physics-based approach to a data-driven approach. To this aim, we present the development of a deep learning model for the re-entry prediction of uncontrolled objects in Low Earth Orbit (LEO). The model is based on a modified version of the Sequence-to-Sequence architecture and is trained on the average altitude profile as derived from a set of Two-Line Element (TLE) data of over 400 bodies. The novelty of the work consists in introducing in the deep learning model, alongside the average altitude, three new input features: a drag-like coefficient ($B^{*}$), the average solar index, and the area-to-mass ratio of the object. 
The developed model is tested on a set of objects studied in the Inter-Agency Space Debris Coordination Committee (IADC) campaigns. The results show that the best performances are obtained on bodies characterised by the same drag-like coefficient and eccentricity distribution as the training set.}


\begin{document}

\section{Introduction}
\label{sec:introduction}

From the launch of the first artificial satellite on the $4^{th}$ of October 1957, the population of bodies orbiting around the Earth has been continuously growing with an exponential trend \cite{esa_report}. Furthermore, in the last years, this growth has been exacerbated by the creation of large constellations for telecommunication, navigation and global internet coverage. Post-mission disposal guidelines  \citep{ALBY20041260, nasa_debris} have been introduced in order to preserve the Low Earth Orbit (LEO) region and incentivise the early disposal of satellites so that they would not affect the already polluted orbital environment. Focusing on the re-entry problem, as of March 2022, a total of 26135 bodies have re-entered the atmosphere\footnote{www.space-track.org}. According to Pardini et al. \citep{results_uncontrolled}, since April 2013, only one out of five re-entries of intact objects was controlled, corresponding to a total mass of roughly 11000 metric tonnes, mainly related to rocket bodies and spacecraft. Therefore, the continuously growing number of orbiting objects should be correlated with an expected increase in the re-entries of both controlled and uncontrolled objects. However, predicting the re-entry epoch and the location of an uncontrolled body remains critical because it is affected by several uncertainties, which could translate into risks of casualties on the ground \cite{results_uncontrolled}. 


In the typical dynamics-based approach to the re-entry prediction problem, the trajectory of the spacecraft is propagated until it reaches the altitude when the break-up occurs. This approach consists of determining the initial position of the object and accurately modelling the forces acting on it in order to predict the evolution of its trajectory. However, in general, it is difficult to accurately model the forces acting on the spacecraft and to precisely know its physical characteristics and state. In fact, a typically recommended relative uncertainty value of $\pm$20\% is recommended for re-entry predictions \citep{pardini2003performance,PARDINI201846}. \cite{results_uncontrolled} give an in-depth review of the different sources of uncertainties in re-entry predictions, the most relevant being the modelling of the drag acceleration. Other forces affect the re-entry dynamics such as solar radiation pressure and the lunisolar attraction. However, in this work, we focus on the latest stage of the re-entry prediction of low eccentricity orbits; therefore, the dominant force is atmospheric drag and the main uncertainties are related to the atmospheric density. It is modelled through empirical models which describe the density variations on spatial and temporal scales \citep{solarmodel,VALLADO2014141,vallado2001fundamentals,anselmo2005computational,FREY20191}. However, they are affected by two types of uncertainties \cite{anselmo2005computational}: the simplified physical modelling on which they are based; and the uncertainties and complexities in predicting the space weather, in particular the solar index. Another important variable for the predictions of low eccentricity orbits is the ballistic coefficient of the object of interest. The ballistic coefficient depends on the object's shape, orientation, and attitude, which can be difficult to accurately know for uncontrolled objects.

In this circumstance, it becomes interesting to adopt a paradigm shift to model the problem, from a physical to a data-driven point of view. In this context, machine learning represents a new approach, where, instead of accurately modelling the space environment, its forces and how the spacecraft interacts with them, it autonomously builds its knowledge based on data, autonomously dealing with the complexities and uncertainties of the phenomenon. Jung et al. \citep{jung2021recurrent} proposed a machine learning approach to predict the re-entry epoch of an uncontrolled satellite based on Recurrent Neural Networks (RNNs). In their work, the RNN is trained on a set of Two-Line Elements (TLEs) sets using as the only feature the evolution of the average altitude of the object. TLE sets provide, among other information, the time evolution of the position of catalogued objects in orbit. However, as also addressed by Lidtke et al. \citep{LIDTKE20191289}, TLEs suffer from some inaccuracies that must be taken into account. Most notably, TLEs of an object are not homogeneous; they can be provided at irregular time steps and contain incorrect observations or observations belonging to other objects (outliers). In addition, TLEs do not directly provide information on the physical characteristics of the object (e.g., the ballistic coefficient), rather, they combine all the uncertainties and modelling errors in a single drag-like coefficient, referred to as $B^{*}$ \citep{VALLADO2014141,vallado2001fundamentals}. Finally, TLEs do not contain accuracy information on the measure they provide that is they do not provide covariance information. However, TLE sets are currently the only public source of data that can be used for re-entry predictions. Therefore, despite their limitations and inaccuracies, they are used in this work to assess the capabilities of a machine-learning approach using multiple input features. As shown in \cite{flohrer2008assessment}, initial covariance information can be estimated using pseudo-observations, and the accuracy of TLE data can be improved as shown \cite{levit2011improved,aida2013accuracy}. While in this work we focus on the application of a machine learning technique to raw TLE data, future work can benefit form the inclusion of such information to increase the accuracy of the re-entry predictions.

Starting from the work of Jung et al. \citep{jung2021recurrent}, we expand on the machine learning model introducing additional features, namely the drag-like coefficient, $B^{*}$, and the solar index. The first feature will allow the machine learning model to account for the physical characteristics of the objects, while the second feature will give a temporal dimension to the model, accounting for the different solar activity levels and how they affect the lifetime of an object. The developed model is trained on hundreds of TLE sets belonging to uncontrolled objects such as spent payloads and rocket bodies, and debris. Consequently, the same model can be tested on different objects, in order to obtain the output sequence and assess its accuracy.

The paper is structured as follows. \cref{sec:deep_model} describes the deep learning model architecture adopted, \cref{sec:data_processing} contains the procedure used for the pre-processing and filtering of the data required by the deep learning model, \cref{sec:features} contains an analysis of the characteristics of the input features of the model, and \cref{sec:results} shows the results obtained by the model on selected test cases, together with the optimisation procedure of the model's hyperparameters.


\section{Deep learning model}
\label{sec:deep_model}
From a machine learning perspective, the task of predicting the re-entry of an object is a time-series problem, where the aim of the model consists in predicting the evolution of the trajectory based on an initial set of conditions so that the re-entry epoch can be determined. In other words, it is a regression problem. Time series are discrete-time data and can be considered as a particular case of sequential data, where each value is associated with a time stamp. Therefore, the order of the series is imposed by the time dimension and it must be preserved. Let us denote an input sequence of length $T_x$ as \cite{raschka2015python}:

\begin{equation}
    \bm{x}=\{\bm{x}^{<1>},\bm{x}^{<2>},...,\bm{x}^{<T_x>}\}
\end{equation}

where the vector $\bm{x}^{<t>} \in \mathbb{R}^{p}$ contains the input features at time step $t$. Furthermore, let us denote a target sequence of length $T_y$ as \cite{raschka2015python}:

\begin{equation}
    \bm{y}=\{{y}^{<1>},{y}^{<2>},...,{y}^{<T_y>}\}
\end{equation}

where $y^{<t>} \in \mathbb{R}$ is the target feature. We define the predicted sequence as \cite{raschka2015python}:

\begin{equation}
    \bm{\hat{y}}=\{{\hat{y}}^{<1>},{\hat{y}}^{<2>},...,{\hat{y}}^{<T_y>}\}
\end{equation}

The aim of the deep learning problem consists in predicting the target sequence, minimising a loss function, that in this work has been selected as the Mean Squared Error (MSE) \cite{raschka2015python}:

\begin{equation}
    \mathcal{L}_{MSE}(\bm{y},\bm{\hat{y}}) =\frac{1}{T_y} \sum_{t=1}^{T_y} \left({y}^{<t>} - {\hat{y}}^{<t>}\right)^{2}
    \label{eq:loss_mse}
\end{equation}

The choice of the MSE is natural in this context since the similarity with the Euclidean distance. Indeed, the MSE can be intuitively considered as the squared distance between the true position and the predicted one, averaged by the number of observations in the output sequence.

\subsection{Sequence-to-Sequence architecture}
\label{subsec:seq2seq}
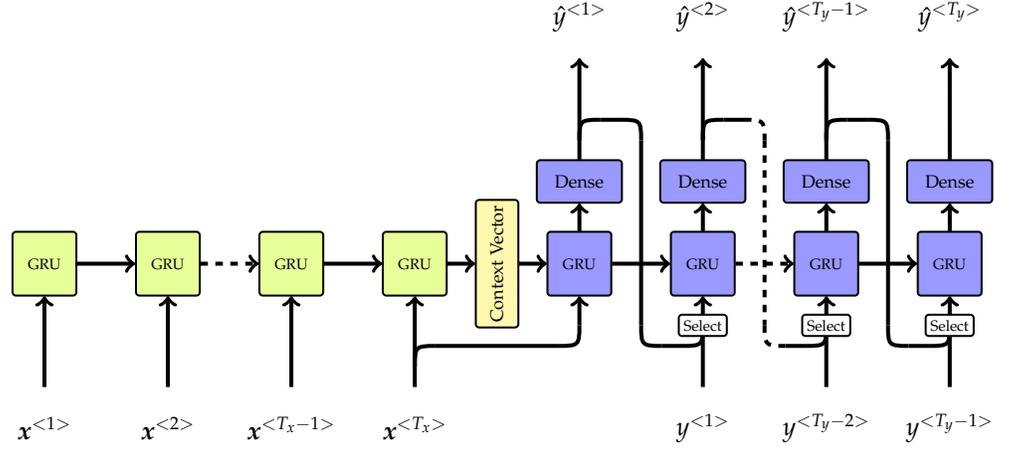
\begin{figure*}[!htb]
\centering
\resizebox{13.2cm}{!}{%
\begin{tikzpicture}[thick,scale=0.58, every node/.style={scale=0.6}]
	\begin{pgfonlayer}{nodelayer}
		\node [shape=rectangle, draw=black, fill=lime!40,  minimum size=1.5cm, thick, rounded corners=1.4] (0) at (0, 0) {GRU};
		\node [shape=rectangle, draw=black, fill=lime!40,  minimum size=1.5cm, thick, rounded corners=1.4] (1) at (3, 0) {GRU};
		\node [shape=rectangle, draw=black, fill=lime!40,  minimum size=1.5cm, thick, rounded corners=1.4] (2) at (6, 0) {GRU};
		\node [shape=rectangle, draw=black, fill=lime!40,  minimum size=1.5cm, thick, rounded corners=1.4] (3) at (9, 0) {GRU};
		\node [shape=rectangle, draw=black, fill=yellow!40,  minimum height=1cm, minimum width=3cm, thick, rounded corners=1.4, rotate=90] (4) at (11, 0) {\large Context Vector};
		\node [shape=rectangle, draw=black, fill=blue!40,  minimum size=1.5cm, thick, rounded corners=1.3] (5) at (13, 0) {GRU};
		\node [shape=rectangle, draw=black, fill=blue!40,  minimum size=1.5cm, thick, rounded corners=1.3] (6) at (16, 0) {GRU};
		\node [shape=rectangle, draw=black, fill=blue!40,  minimum size=1.5cm, thick, rounded corners=1.3] (7) at (19, 0) {GRU};
		\node [shape=rectangle, draw=black, fill=blue!40,  minimum size=1.5cm, thick, rounded corners=1.3] (8) at (22, 0) {GRU};
		\node [] (9) at (0, -3) {};
		\node [] (10) at (3, -3) {};
		\node [] (11) at (6, -3) {};
		\node [] (12) at (9, -3) {};
		\node [shape=rectangle, draw=black, fill=blue!40,  minimum height=1cm, minimum width=2cm, thick, rounded corners=1.4] (13) at (13, 2) {\large Dense};
		\node [shape=rectangle, draw=black, fill=blue!40,  minimum height=1cm, minimum width=2cm, thick, rounded corners=1.4] (14) at (16, 2) {\large Dense};
		\node [shape=rectangle, draw=black, fill=blue!40,  minimum height=1cm, minimum width=2cm, thick, rounded corners=1.4] (15) at (19, 2) {\large Dense};
		\node [shape=rectangle, draw=black, fill=blue!40,  minimum height=1cm, minimum width=2cm, thick, rounded corners=1.4] (16) at (22, 2) {\large Dense};
		\node [] (17) at (13, 5) {};
		\node [] (18) at (16, 5) {};
		\node [] (19) at (19, 5) {};
		\node [] (20) at (22, 5) {};
		\node [] (21) at (13, 3) {};
		\node [] (24) at (22, 3.5) {};
		\node [] (25) at (14, 3.5) {};
		\node [] (26) at (14.5, 3) {};
		\node [] (27) at (14.5, -1.5) {};
		\node [] (28) at (15, -2) {};
		\node [] (29) at (15.5, -2) {};
		\node [shape=rectangle, draw=black, fill=white,  minimum size=.5cm, thick, rounded corners=1.3] (30) at (16, -1.5) {Select};
		\node [] (31) at (13.5, 3.5) {};
		\node [] (32) at (16, 3) {};
		\node [] (33) at (17, 3.5) {};
		\node [] (34) at (17.5, 3) {};
		\node [] (35) at (17.5, -1.5) {};
		\node [] (36) at (18, -2) {};
		\node [] (37) at (18.5, -2) {};
		\node [shape=rectangle, draw=black, fill=white,  minimum size=.5cm, thick, rounded corners=1.3] (38) at (19, -1.5) {Select};
		\node [] (39) at (16.5, 3.5) {};
		\node [] (40) at (9, -2.5) {};
		\node [] (41) at (9.5, -2) {};
		\node [] (42) at (13, -1.5) {};
		\node [] (43) at (12.5, -2) {};
		\node [] (44) at (19, 3) {};
		\node [] (45) at (20, 3.5) {};
		\node [] (46) at (20.5, 3) {};
		\node [] (47) at (20.5, -1.5) {};
		\node [] (48) at (21, -2) {};
		\node [] (49) at (21.5, -2) {};
		\node [shape=rectangle, draw=black, fill=white,  minimum size=.5cm, thick, rounded corners=1.3] (50) at (22, -1.5) {Select};
		\node [] (51) at (19.5, 3.5) {};
		\node [] (52) at (0, -4) {\LARGE $\bm{x}^{<1>}$};
		\node [] (53) at (3, -4) {\LARGE $\bm{x}^{<2>}$};
		\node [] (54) at (6, -4) {\LARGE $\bm{x}^{<T_x-1>}$};
		\node [] (55) at (9, -4) {\LARGE $\bm{x}^{<T_x>}$};
		\node [] (56) at (13, 6) {\LARGE $\hat{y}^{<1>}$};
		\node [] (57) at (16, 6) {\LARGE $\hat{y}^{<2>}$};
		\node [] (58) at (19, 6) {\LARGE $\hat{y}^{<T_y-1>}$};
		\node [] (59) at (22, 6) {\LARGE $\hat{y}^{<T_y>}$};
		\node [] (60) at (16, -3) {};
		\node [] (61) at (19, -3) {};
		\node [] (62) at (22, -3) {};
		\node [] (63) at (16, -4) {\LARGE ${y}^{<1>}$};
		\node [] (64) at (19, -4) {\LARGE ${y}^{<T_y -2>}$};
		\node [] (65) at (22, -4) {\LARGE ${y}^{<T_y-1>}$};
	\end{pgfonlayer}
	\begin{pgfonlayer}{edgelayer}
		\draw [ultra thick, ->] (9.center) to (0);
		\draw [ultra thick, ->] (10.center) to (1);
		\draw [ultra thick, ->] (11.center) to (2);
		\draw [ultra thick, ->] (12.center) to (3);
		\draw [ultra thick, ->] (0) to (1);
		\draw [ultra thick, ->, dashed] (1) to (2);
		\draw [ultra thick, ->] (2) to (3);
		\draw [ultra thick, ->] (3) to (4);
		\draw [ultra thick, ->] (4) to (5);
		\draw [ultra thick, ->] (5) to (6);
		\draw [ultra thick, ->, dashed] (6) to (7);
		\draw [ultra thick, ->] (7) to (8);
		\draw [ultra thick, ->] (13) to (17.center);
		\draw [ultra thick, ->] (14) to (18.center);
		\draw [ultra thick, ->] (15) to (19.center);
		\draw [ultra thick, ->] (16) to (20.center);
		\draw [ultra thick, ->] (5) to (13);
		\draw [ultra thick, ->] (6) to (14);
		\draw [ultra thick, ->] (7) to (15);
		\draw [ultra thick, ->] (8) to (16);
		\draw [bend right=45, looseness=1.75, ultra thick] (27.center) to (28.center);
		\draw [ultra thick] (28.center) to (29.center);
		\draw [bend right=45, looseness=1.75, ultra thick] (29.center) to (30.center);
		\draw [ultra thick, ->] (30.center) to (6);
		\draw [bend left=45, looseness=1.75, ultra thick] (21.center) to (31.center);
		\draw [ultra thick] (31.center) to (25.center);
		\draw [bend left=45, looseness=1.75, ultra thick] (25.center) to (26.center);
		\draw [ultra thick] (26.center) to (27.center);
		\draw [bend right=45, looseness=1.75, ultra thick, dashed] (35.center) to (36.center);
		\draw [ultra thick] (36.center) to (37.center);
		\draw [bend right=45, looseness=1.75, ultra thick] (37.center) to (38.center);
		\draw [bend left=45, looseness=1.75, ultra thick] (32.center) to (39.center);
		\draw [ultra thick] (39.center) to (33.center);
		\draw [bend left=45, looseness=1.75, ultra thick, dashed] (33.center) to (34.center);
		\draw [ultra thick, dashed] (34.center) to (35.center);
		\draw [ultra thick, ->] (38.center) to (7);
		\draw [bend left=45, looseness=1.75, ultra thick] (40.center) to (41.center);
		\draw [ultra thick] (41.center) to (43.center);
		\draw [bend right=45, looseness=1.75, ultra thick] (43.center) to (42.center);
		\draw [ultra thick, ->] (42.center) to (5);
		\draw [bend right=45, looseness=1.75, ultra thick] (47.center) to (48.center);
		\draw [ultra thick] (48.center) to (49.center);
		\draw [bend right=45, looseness=1.75, ultra thick] (49.center) to (50.center);
		\draw [bend left=45, looseness=1.75, ultra thick] (44.center) to (51.center);
		\draw [ultra thick] (51.center) to (45.center);
		\draw [bend left=45, looseness=1.75, ultra thick] (45.center) to (46.center);
		\draw [ultra thick] (46.center) to (47.center);
		\draw [ultra thick, ->] (50.center) to (8);
		\draw [ultra thick] (30.center) to (60.center);
		\draw [ultra thick] (38.center) to (61.center);
		\draw [ultra thick] (50.center) to (62.center);
	\end{pgfonlayer}
\end{tikzpicture}
}
\caption{Representation of Seq2Seq architecture during training.}\label{fig:seq2seq_complete}
\end{figure*}

In this work, we selected a Sequence-to-Sequence (Seq2Seq) \cite{sutskever2014seq2seq} architecture to model the re-entry problem, due to its ability to deal with input and output sequences of different lengths. As shown in \cref{fig:seq2seq_complete} , a Seq2Seq architecture is composed of two neural networks, the encoder and the decoder, and the context vector. The encoder receives as input a sequence of arbitrary length, $T_x$, and generates a fixed-length vector, called the context vector. This vector summarises the information contained in the input sequence. The outputs of the decoder are simply discarded. At each time step, the cell produces the hidden state considering the input at the same time step and the hidden state at the previous time step \cite{zhang2021dive}:

\begin{equation}
    \bm{h}_{Enc}^{<t>} = f(\bm{x}^{<t>}, \bm{h}_{Enc}^{<t-1>}, \bm{\theta})
\end{equation}

where $\bm{h}_{Enc}^{<t>}$ is the hidden state of the encoder at time $t$, $\bm{h}_{Enc}^{<t-1>}$ is the hidden state at the previous time step, $\bm{x}^{<t>}$ is the input vector at the current time, and $\bm{\theta}$ is the weights vector. The function $f$ represents a neural network, which can be a simple Recurrent Neural Network (RNN) \citep{goodfellow2016deep,elman1990finding} or a more complex Gate Recurrent Unit (GRU) \citep{goodfellow2016deep,cho-etal-2014-learning,zhang2021dive} or a Long-Short Term Memory (LSTM) \citep{goodfellow2016deep,zhang2021dive} network. At the final time step, $T_x$, the hidden state of the encoder is defined as the context vector as follows:

\begin{equation}
    \bm{c}_{context} = \bm{h}_{Enc}^{<T_x>}
\end{equation}

The decoder is initialised with the final state of the encoder, $\bm{h}_{Dec}^{<t'=0>} = \bm{c}_{context}$. The aim of the decoder is to generate the output sequence based on this encoded vector. At each time step $t' \in [1, T_y]$, given the hidden state and the output at the previous time step $t'-1$, the decoder hidden state can be defined as follows \citep{zhang2021dive}:

\begin{equation}
    \bm{h}_{Dec}^{<t'>} = g(\hat{y}^{<t'-1>}, \bm{h}_{Dec}^{<t'-1>}, \bm{\theta}')
\end{equation}

where again $g(\cdot)$ can be a simple RNN, a GRU or an LSTM. Subsequently, it is possible to obtain the output $\hat{y}^{<t'>}$ by passing the corresponding output of the decoder to an output layer. Note that the hidden state at time $t'=T_y$ is simply discarded.
\bigbreak
During training, the decoder receives its own previous hidden state and the true input, or the predicted output at the previous time step, to generate a new state and output sequentially. When only the "true input" is provided to the decoder, the training method is referred to as Teacher Forcing \citep{goodfellow2016deep}. Therefore, in this method, the predicted output of the model is substituted with our reference data, which is considered the ground truth. However, this method introduces a discrepancy between the training phase and the inference phase in which the network is used in open-loop, which can introduce errors in the prediction as the network is not able to learn from its own mistakes. To mitigate this problem, Bengio et al. \cite{bengio2015scheduled} proposed an approach called Curriculum Learning. It consists of randomly selecting either the previous true output or the model estimate, with a probability that is proportional to the training epoch. Intuitively, at the initial epochs of training, the model should be trained with teacher forcing since it is not still well trained and errors can be made. Instead at the final epochs, the model is expected to be well trained so that it is possible to use the model output, as during inference. Specifically, for every output to predict $y^{<t'>}$ of every batch at the \textit{j-th} epoch, a coin is flipped to decide to feed either the previous true output, associated with a probability $\epsilon_j$, or the model estimate, with a probability $(1 - \epsilon_j)$. Given the fact that in this context the size of the dataset is quite small and the training epoch number is large, the probability is set as a function of the training epoch, instead of the iteration \cite{bengio2015scheduled}, as:

\begin{equation}
    \epsilon_j = k^j
    \label{eq:decay_SS}
\end{equation}

where $k<1$ is a constant that depends on the speed of convergence, which is the decay rate of the scheduled sampling.
\bigbreak
The representation of the Seq2Seq model during training is provided in \cref{fig:seq2seq_complete}, where $\bm{x}^{<t>}$ denotes each time component of the input sequence; $\hat{y}^{<t>}$ the predicted output; and $y^{<t>}$ the ground truth.

\subsubsection{Gate Recurrent Unit}
\label{subsubsec;gru}
In this work, both the encoder and the decoder are based on Gate Recurrent Units (GRUs) \citep{cho-etal-2014-learning}, given their capability of dealing with the vanishing gradient problem that characterises simple RNN when trained with long sequences. It can be seen as a simplified version of the LSTM and it has the advantage to reduce the computational cost, providing comparable performances \cite{chung2014empirical}. At each time step, the GRU computes:

\begin{align}
\mathbf{z}^{<t>} &=\sigma\left(\mathbf{W}_{z} \mathbf{x}^{<t>}+\mathbf{V}_{z} \mathbf{c}^{<t-1>}+\mathbf{b}_{z}\right) \label{eq:update_gate}\\
\mathbf{r}^{<t>} &=\sigma\left(\mathbf{W}_{r} \mathbf{x}^{<t>}+\mathbf{V}_{r} \mathbf{c}^{<t-1>}+\mathbf{b}_{r}\right) \label{eq:reset_gate}  \\
\tilde{\mathbf{c}}^{<t>} &=\tanh \left(\mathbf{W}_{c} \mathbf{x}_{t}+\mathbf{V}_{c}\left(\mathbf{r}^{<t>} \odot \mathbf{c}^{<t-1>}\right) + \mathbf{b}_c\right) \label{eq:candidate}  \\
\mathbf{c}^{<t>} &=\left(1-\mathbf{z}^{<t>}\right) \odot \tilde{\mathbf{c}}^{<t>}+\mathbf{z}^{<t>} \odot {\mathbf{c}}^{<t-1>} \label{eq:memory_cell}
\end{align}

where $\bm{W}_z$, $\bm{W}_r$, $\bm{W}_c \in \mathbb{R}^{k \times p}$ denote the input weight matrices; $\bm{V}_z$, $\bm{V}_r$, $\bm{V}_c \in \mathbb{R}^{k \times k}$ are the hidden state weights; $\bm{b}_z$, $\bm{b}_r$, $\bm{b}_c \in \mathbb{R}^{k}$ are the bias vectors; $ \odot$ the element-wise product; $\sigma(\cdot)$ and $\tanh(\cdot)$ are the sigmoid function and the hyperbolic tangent function, respectively. The variables $\bm{z}^{<t>} \in \mathbb{R}^{k}$ and $\bm{r}^{<t>} \in \mathbb{R}^{k}$ represent the update and reset gate respectively, which helps to tackle the vanishing gradient problem. Their aim consists in controlling the quantity of information of the past time step that needs to be carried forward. In particular, the update gate is responsible for updating the old state, by controlling the similarity between the new state and the past one. The reset gate is responsible for setting the amount of information that needs to be forgotten. For these reasons, the values of the gates are designed to be bounded between 0 and 1, by using the sigmoid function. The output hidden state is computed in two steps. First, the candidate memory cell $\tilde{\bm{c}} \in \mathbb{R}^{k}$ is calculated according to \cref{eq:candidate}. It takes as input the previous hidden state, which is multiplied by the reset gate, and the current input. Depending on the entries of the reset gate, two limit cases can be identified. When its values are close to unity, the information of the previous time step is carried forward in time, like a simple RNN. Instead, when the values are close to 0, the information from the previous time step is ignored.
After having derived the candidate memory cell, the variable $\bm{c} \in \mathbb{R}^{k}$, called memory cell, can be computed according to \cref{eq:memory_cell}. The memory cell receives as input the current candidate and the previous memory cell and it is governed by the reset gate only. In this case, when the entries of the reset gate are close to 0, the new memory cell is updated with the candidate, meaning that the information of the previous time step is dropped. Instead, when the values of the gate approach to unity, the candidate is simply discarded and the memory cell maintains its state. Finally, the hidden state and the output of the GRU can be defined as, respectively:

\begin{align}
    \bm{h}^{<t>} &= \bm{c}_{context}^{<t>}\\
    \bm{\hat{o}}^{<t>} &= \bm{c}_{context}^{<t>} 
\end{align}

\begin{figure}[!htb]
    \centering
   \begin{tikzpicture}[scale=0.45]
	\begin{pgfonlayer}{nodelayer}
		\node [] (11) at (-12, 4) {};
		\node [] (12) at (-7.5, 4) {};
		\node [] (13) at (-7, -8) {};
		\node [] (15) at (-1.5, -3) {};
		\node [] (17) at (-2.5, 0.5) {};
		\node [shape=circle, draw=black, fill=blue!40,  minimum size=0.7cm] (19) at (-2, -1.5) {$\sigma$};
		\node [shape=circle, draw=black, fill=blue!40,  minimum size=0.7cm] (20) at (2, 4) {$\odot$};
		\node [shape=circle, draw=black, fill=blue!40,  minimum size=0.7cm] (21) at (6, 4) {$+$};
		\node [] (22) at (12, 4) {};
		\node [] (23) at (-5, 3.5) {};
		\node [shape=circle, draw=black, fill=blue!40,  minimum size=0.7cm] (24) at (-5, 0.5) {$\odot$};
		\node [] (25) at (-6.5, -5) {};
		\node [] (26) at (6, -4.5) {};
		\node [] (27) at (-5, 0.5) {};
		\node [shape=circle, draw=black, fill=blue!40,  minimum size=0.7cm] (28) at (6, 0.5) {$\odot$};
		\node [] (29) at (-5, -4.5) {};
		\node [] (30) at (2, -2.5) {};
		\node [shape=circle, draw=black, fill=blue!40,  minimum size=0.7cm] (33) at (2, -1.5) {$\sigma$};
		\node [shape=circle, draw=black, fill=blue!40,  minimum size=0.7cm] (35) at (6, -1.5) {$\scriptscriptstyle tanh$};
		\node [] (36) at (-2, 0) {};
		\node [] (37) at (-7, -4.5) {};
		\node [] (38) at (-7, -5.5) {};
		\node [] (40) at (-7, -3.5) {};
		\node [] (42) at (-4.5, -5) {};
		\node [] (43) at (-6.5, -3) {};
		\node [] (44) at (-7, -2.5) {};
		\node [] (45) at (-7, 3.5) {};
		\node [] (46) at (-5.5, 4) {};
		\node [] (47) at (-2.5, -3) {};
		\node [] (48) at (-2, -2.5) {};
		\node [] (49) at (1.5, -3) {};
		\node [] (50) at (2, 1) {};
		\node [] (51) at (2, 0) {};
		\node [] (52) at (2.5, 0.5) {};
		\node [] (53) at (-4.5, 4) {};
		\node [] (54) at (5.5, -5) {};
		\node [] (55) at (-1.5, 5) {};
		\node [] (56) at (-5.5, 5) {};
		\node [] (57) at (-6, -3.25) {};
		\node [] (58) at (-1.5, -3.75) {};
		\node [] (59) at (-12, 5) {$\bm{c}^{<t-1>} = \bm{h}^{<t-1>}$};
		\node [] (60) at (12, 5) {$\bm{c}^{<t>} = \bm{h}^{<t>}$};
		\node [] (61) at (-7, -9) {$\bm{x}^{<t>}$};
		\node [] (62) at (-6, 4.5) {};
		\node [] (63) at (-1, 4.5) {};
		\node [] (64) at (-5.5, -3.75) {};
		\node [] (65) at (-1, -3.25) {};
		\node [] (66) at (1, 5) {};
		\node [] (67) at (0.5, 4.5) {};
		\node [] (68) at (3, 5) {};
		\node [] (69) at (3.5, 4.5) {};
		\node [] (70) at (0.5, -3.25) {};
		\node [] (71) at (1, -3.75) {};
		\node [] (72) at (3, -3.75) {};
		\node [] (73) at (3.5, -3.25) {};
		\node [] (74) at (2, 5) {};
		\node [] (75) at (2, 6.5) {};
		\node [] (76) at (2, 7) {Update gate};
		\node [] (77) at (-3.5, 5) {};
		\node [] (78) at (-3.5, 6.5) {};
		\node [] (79) at (-3.5, 7) {Reset gate};
		\node [shape=circle, draw=black, fill=blue!40,  minimum size=0.7cm] (81) at (4, 0.5) {$\scriptscriptstyle 1-$}; 
		\draw [black, very thick, shape=rounded rectangle] (-9,-6) rectangle (9,6);
	\end{pgfonlayer}
	\begin{pgfonlayer}{edgelayer}
		\draw [ultra thick, ->] (20) to (21);
		\draw [ultra thick, ->] (21) to (22.center);
		\draw [ultra thick, ->] (23.center) to (24);
		\draw [in=270, out=90, ultra thick ,->] (28) to (21);
		\draw [ultra thick] (27.center) to (29.center);
		\draw [ultra thick] (26.center) to (35);
		\draw [ultra thick, ->] (35) to (28);
		\draw [ultra thick, , <-] (27.center) to (17.center);
		\draw [bend left=45, looseness=1.50, ultra thick] (17.center) to (36.center);
		\draw [ultra thick] (36.center) to (19);
		\draw [ultra thick, ->] (30.center) to (33);
		\draw [ultra thick] (40.center) to (37.center);
		\draw [ultra thick] (38.center) to (13.center);
		\draw [bend right=45, looseness=1.50, ultra thick] (25.center) to (38.center);
		\draw [bend right=45, looseness=1.75, ultra thick] (29.center) to (42.center);
		\draw [bend left=45, looseness=1.50, ultra thick] (40.center) to (43.center);
		\draw [bend left=45, looseness=1.75, ultra thick] (43.center) to (44.center);
		\draw [ultra thick] (12.center) to (46.center);
		\draw [ultra thick] (44.center) to (45.center);
		\draw [ultra thick] (38.center) to (37.center);
		\draw [ultra thick] (43.center) to (47.center);
		\draw [bend right=45, looseness=1.75, ultra thick] (47.center) to (48.center);
		\draw [ultra thick] (47.center) to (15.center);
		\draw [ultra thick, <-] (19) to (48.center);
		\draw [bend right=45, looseness=1.75, ultra thick] (49.center) to (30.center);
		\draw [ultra thick] (15.center) to (49.center);
		\draw [ultra thick] (33) to (51.center);
		\draw [ultra thick] (51.center) to (50.center);
		\draw [bend left=45, looseness=1.75, ultra thick] (51.center) to (52.center);
		\draw [ultra thick, ->] (52.center) to (81);
		\draw [ultra thick, ->] (81) to (28);
		\draw [ultra thick] (11.center) to (12.center);
		\draw [ultra thick, ->] (50.center) to (20);
		\draw [ultra thick] (46.center) to (53.center);
		\draw [ultra thick, ->](53.center) to (20);
		\draw [ultra thick] (25.center) to (42.center);
		\draw [ultra thick] (42.center) to (54.center);
		\draw [bend right=45, looseness=1.75, ultra thick] (54.center) to (26.center);
		\draw [bend left=45, looseness=1.75, ultra thick] (12.center) to (45.center);
		\draw [bend left=45, looseness=1.75, ultra thick] (46.center) to (23.center);
		\draw [bend left=45, looseness=1.75, dashed] (62.center) to (56.center);
		\draw [dashed] (56.center) to (55.center);
		\draw [bend left=45, looseness=1.75, dashed] (55.center) to (63.center);
		\draw [bend right=45, looseness=1.75, dashed] (57.center) to (64.center);
		\draw [dashed] (64.center) to (58.center);
		\draw [bend right=45, looseness=1.75, dashed] (58.center) to (65.center);
		\draw [dashed] (63.center) to (65.center);
		\draw [dashed] (62.center) to (57.center);
		\draw [dashed] (66.center) to (68.center);
		\draw [bend left=45, looseness=1.75, dashed] (68.center) to (69.center);
		\draw [bend right=45, looseness=1.75, dashed] (66.center) to (67.center);
		\draw [bend right=315, looseness=1.75, dashed] (71.center) to (70.center);
		\draw [dashed] (67.center) to (70.center);
		\draw [dashed] (69.center) to (73.center);
		\draw [bend left=45, looseness=1.75, dashed] (73.center) to (72.center);
		\draw [dashed] (72.center) to (71.center);
		\draw [ thick] (74.center) to (75.center);
		\draw [ thick] (77.center) to (78.center);
	\end{pgfonlayer}
\end{tikzpicture}
\caption{Computational flow of GRU architecture.} \label{fig:GRU}
\end{figure}
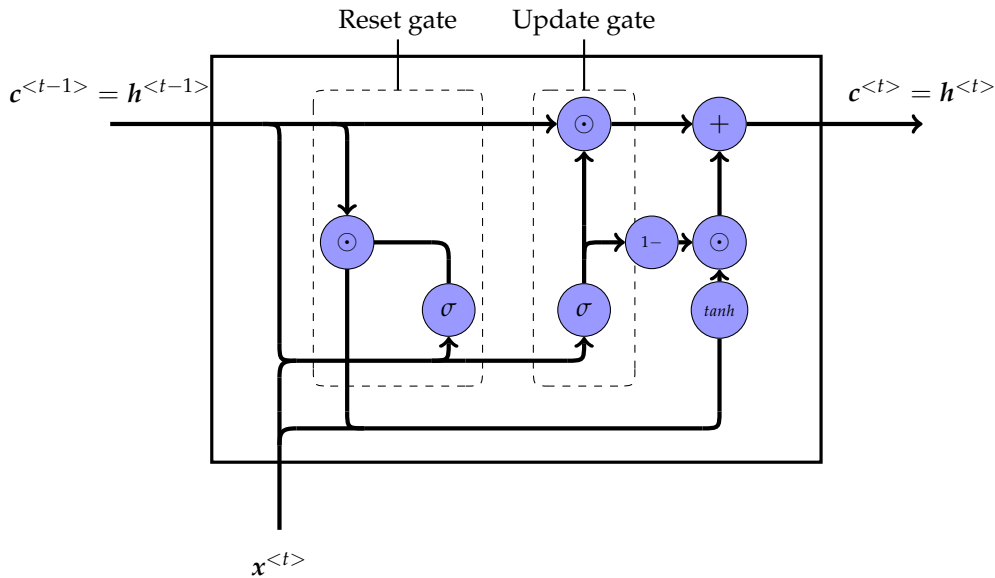

The computational flow diagram of a GRU is summarised in \cref{fig:GRU}.

\bigbreak
The Seq2Seq model is built on TensorFlow \cite{tensorflow2015-whitepaper}, a widely used open-source library for machine and deep learning, developed by Google. TensorFlow automatically derives the gradients with respect to the weights of the model using auto-differentiation, making the training of the model a straightforward task. The high-level interface to TensorFlow is provided by Keras \cite{chollet2021deep}, a \textit{Python} library that allows building, training, and testing neural networks.


\section{Data pre-processing}
\label{sec:data_processing}
The raw data used to train and validate the deep learning model has been retrieved from Space Track\footnotemark[1]. Specifically, all the decayed objects from the $1^{st}$ of January 2000 to the $7^{th}$ of October 2021, for which the Tracking and Impact Prediction (TIP) is available, have been considered. The TLE data is retrieved in the form of Orbital Mean-Elements Messages (OMM); once the raw TLE data is available, it is necessary to control the quality of the data and prune possible outliers. Following \cite{LIDTKE20191289}, the outliers can be found by focusing on the mean motion and $B^*$ coefficient. The main steps of the pruning procedure are as follows:

\begin{enumerate}
    \item Identify the presence of TLE corrections by defining a time threshold such that the previous observation is considered incorrect. A common time threshold is half an orbit.
    \item Identify large time intervals between consecutive TLEs, and, consequently, divide the entire set into different windows in order to ensure TLEs consistency.
    \item Find and filter out single TLEs with discordant values of the mean motion, based on a robust linear regression technique and on a sliding window of fixed length. In particular, the results obtained through regression are compared with the OMM following the sliding window and an outlier is identified if a relative and an absolute tolerance are exceeded (optimal tuning parameters for the filter can be found in Table 3 of \cite{LIDTKE20191289}).
    \item Identify and filter out possible outliers in eccentricity and inclination, using a statistical approach. Particularly, the mean is computed within a sliding window, and it is subtracted from the central orbital element of the interval, obtaining a time series of differences. Afterwards, using another sliding window that scans the aforementioned series of differences, the mean absolute deviation is computed. An outlier is removed if the orbital element presents a difference from the mean that is higher than a given threshold ((optimal tuning parameters for the filter can be found in Table 4 and Table 5 of \cite{LIDTKE20191289}).
    \item Filter out TLEs that present negative values of $B^*$ because they can be associated with unmodelled dynamics or manoeuvres.
\end{enumerate}


For a complete description of the filters and their performance, as well as the optimal tuning of the filter parameters, the reader is referred to the work of Lidtke et al. \citep{LIDTKE20191289}.
It is worth noting that some outliers may not be identified by such filters; however, the percentage of outliers is considered to be sufficiently small to ensure the validity of the data. After the pruning process, the mean elements contained in the TLEs are converted to the corresponding osculating variables using the SGP4 propagator \citep{vallado2001fundamentals}. It is also important to consider other characteristics of the dataset. Specifically, we want to focus on re-entries that have a sufficient number of TLE points, whose re-entry prediction is sufficiently accurate. The minimum number of TLE points has been selected via a trade-off between the number of pruned TLEs and the accuracy of the fitting procedure (see \cref{subsec:hmean}) \citep{Salmaso2022thesis}. In addition, the orbit should not be too eccentric as the re-entry mechanism for these types of orbits changes significantly. The following criteria have been used to filter the dataset:

\begin{itemize}
    \item Maximum re-entry uncertainty: 20 min.
    \item Maximum average altitude of the initial TLE: 200 km.
    \item Minimum average altitude of the final TLE: 180 km.
    \item Maximum eccentricity: 0.1.
    \item Minimum number of points: 4 TLE.
\end{itemize}

The described criteria adopt of the average altitude, $h$, that is computed according to:

\begin{equation}
    h = a - R_{\oplus},
\end{equation}

where $a$ is the osculating semi major axis; and $R_{\oplus}$ the Earth radius. The result of this pruning and filtering procedure is a dataset with a total of 417 bodies, which is then split in 80\% training and 20\% validation (see \cref{ch:appendix_a}. Note that the objects in the validation set must resemble the characteristics of the training set. For example, the average, $\mu$, and the relative standard deviation, $\sigma$, of the $B^*$ coefficient and the eccentricity for the training and validation sets should be comparable (\cref{tab:means_ecc_bstar}).

\begin{table}[!htb]
\centering
\begin{tabular}{lccc}
\hline
\hline
\multicolumn{2}{c}{\textbf{Variable}}                    & \multicolumn{1}{l}{\textbf{Training Set}} & \multicolumn{1}{l}{\textbf{Validation Set}} \\ \hline
\multirow{2}{*}{\textbf{Ecentricity {[}-{]}}} & $\mu$    & $7.2842 \cdot 10^{-3}$                    & $5.8856 \cdot 10^{-3}$                      \\ \cline{2-4} 
                                              & $\sigma$ & $9.3963 \cdot 10^{-3}$                    & $7.1684 \cdot 10^{-3}$                      \\ \hline
\multirow{2}{*}{\textbf{\boldsymbol{$B^*$} {[}1/ER{]}}}    & $\mu$    & $6.3362\cdot 10^{-4}$                     & $5.9503 \cdot 10^{-4}$                      \\ \cline{2-4} 
                                              & $\sigma$ & $5.4924\cdot 10^{-4}$                     & $3.8341\cdot 10^{-4}$                       \\ \cline{2-4} 
\hline
\hline
\end{tabular}
\caption{$B^*$ and eccentricity characteristics for the training and validation sets.}
\label{tab:means_ecc_bstar}
\end{table}

\begin{figure}[h]
\centering
    \includegraphics[width=0.8\columnwidth]{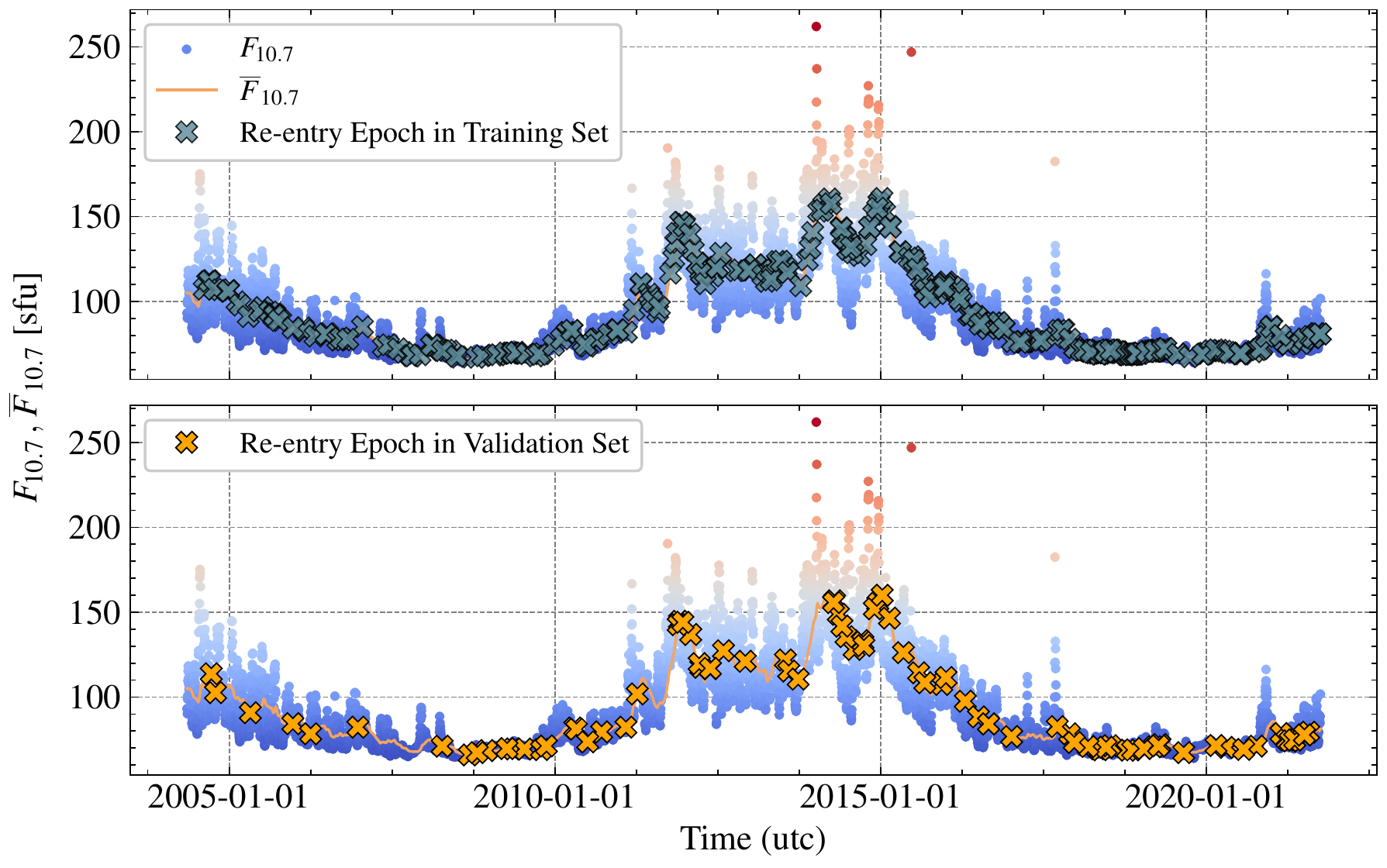}
    \caption{Re-entry epochs in training (Top) and validation (Bottom) set and Solar Flux indexes.}
    \label{fig:epochs_datasets}
\end{figure}

Moreover, the re-entry epochs are distributed quite uniformly from roughly January 2005 to April 2021 for both the sets, therefore the different objects experience different solar flux conditions and, therefore, similar values of drag acceleration (\cref{fig:epochs_datasets}). Finally, it should be highlighted that the majority of the bodies in the training and validation sets is characterised by an uncertainty in the re-entry epoch of less than 2 minutes.\\

The objects for testing the prediction capabilities of the model are selected among the bodies analysed during IADC campaigns, as listed in \cref{tab:test_set}. Note that the re-entry epochs of 23853 and 19045 are manually changed to, respectively, 16:25 UTC and 15:08 UTC \cite{PARDINI201846}.

\begin{table}[!htb]
\centering
\begin{threeparttable}
\begin{tabular}{c c c}
\hline 
\hline
\textbf{NORAD} & \textbf{Name}  & \textbf{Re-entry Epoch} [UTC]\\
\hline
23853 & COSMOS 2332 & 2005-01-28T18:05* \\
\hline
26873 & CORONAS F & 2005-12-06T17:24 \\
\hline
10973 & COSMOS 1025 & 2007-03-10T12:56 \\
\hline
31599 & DELTA 2 R/B & 2007-08-16T09:23 \\
\hline
31928 & EAS & 2008-11-03T04:51 \\
\hline
20813 & MOLNIYA 3-39 & 2009-07-08T22:42\\
\hline
11601 & SL-3 R/B & 2010-04-30T16:44 \\
\hline
21701 & UARS & 2011-09-24T04:00 \\
\hline
20638 & ROSAT & 2011-10-23T01:50 \\
\hline
37872 & PHOBOS-GRUNT & 2012-01-15T17:46 \\
\hline
19045 & COSMOS 1939 & 2014-10-29T15:32*  \\
\hline
40138 & CZ-2D R/B & 2015-06-13T23:58 \\
\hline
39000 & CZ-2C R/B & 2016-06-27T19:04 \\
\hline
38086 & AVUM & 2016-11-02T04:43\\
\hline
37820 &  TIANGONG 1 & 2018-04-02T00:16\\
\hline 
\hline
\end{tabular}

\begin{tablenotes}
      \small
      \item \textit{Note}: the epochs denoted with * are manually changed according to the indications of the IADC campaigns.
    \end{tablenotes}
\caption{List of test objects derived from TIP.}
\label{tab:test_set}
\end{threeparttable}
\end{table}




\section{Features engineering}
\label{sec:features}
In this section, we outline the processes to select and pre-process the most important input features required to predict the re-entry epoch by using a machine learning model. However, this is not a straightforward task because it is necessary to find input variables that are correlated with the output of the deep learning model. Therefore, it is essential to analyse each feature from a physical point of view, so that this knowledge can be then applied to constructing the deep learning model, which is then going to benefit from each input variable. This process is typically known as feature engineering. In this optic, the deep learning model is considered a physical model, where, given the input variables, the output is obtained by means of physical relations. Instead, in machine learning, the model is treated as a black box, which is autonomously capable of learning the input-output relations, based on the true values and the input features. Therefore, each feature can be selected and generated based on the knowledge of the physical principles that govern the re-entry phenomenon, leaving the deep learning model to cope with the learning, and the uncertainties of the physical problem. By analysing each feature, the relations between the considered variables can be highlighted so that we can make a more informed decision on whether the considered feature can be beneficial to the deep learning model \citep{dong2018feature,Salmaso2022thesis}.

In addition, the raw data, provided by the TLEs requires a pre-processing and regularisation step in order for it to be used with an RNN. In fact, TLE data is characterised by a non-uniform generation frequency, which differs from object to object; it also lacks high accuracy and may present outliers. All these aspects must be taken into account before feeding the data to the RNN.
The following sections analyse the contribution of the features considered in this work, which are the average spacecraft altitude, the ballistic coefficient, the solar index, and the $B^*$ coefficient.

\subsection{Average altitude}
\label{subsec:hmean}
\begin{figure}[htb!]
	\centering
	\begin{subfigure}[b]{0.45\textwidth}
        \includegraphics[width=\textwidth]{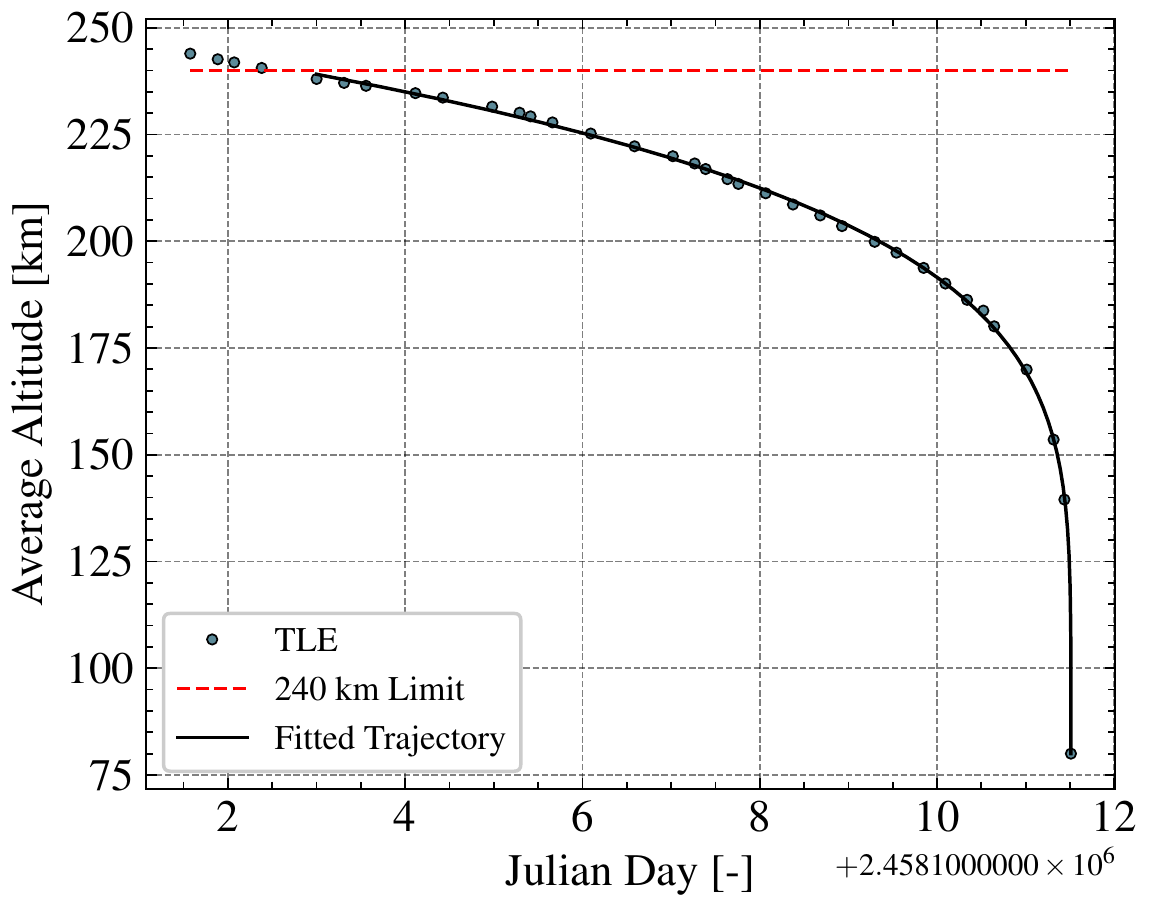}
        \caption{}
	\end{subfigure}
	\hfill
	\begin{subfigure}[b]{0.45\textwidth}
		\centering
		\includegraphics[width=\textwidth]{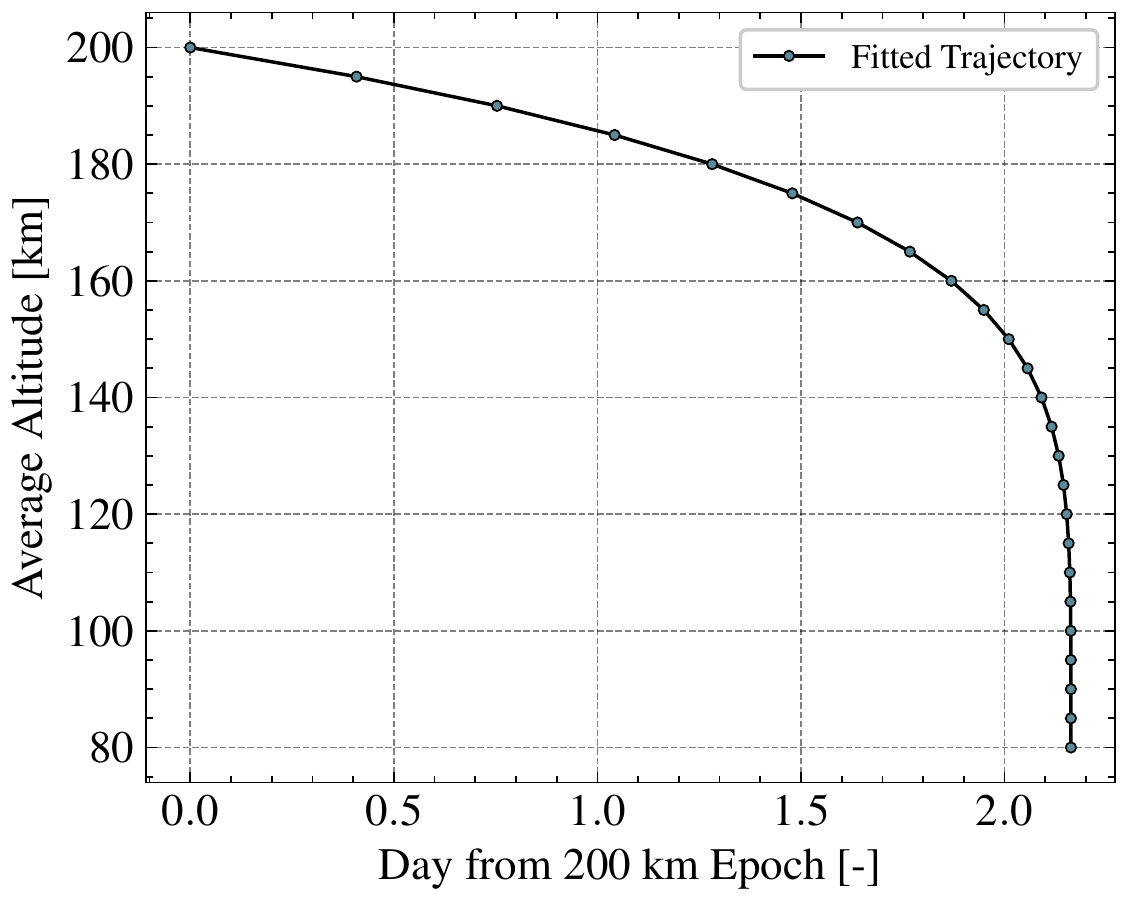}
		\caption{ }
	\end{subfigure}
	\caption{Example of fitted (a) and sampled (b) re-entry trajectory for the object with NORAD number 27392.}
    \label{fig:fit_sample_traj}
\end{figure}

\begin{figure}[htb!]
\centering
    \includegraphics[width=0.9\columnwidth]{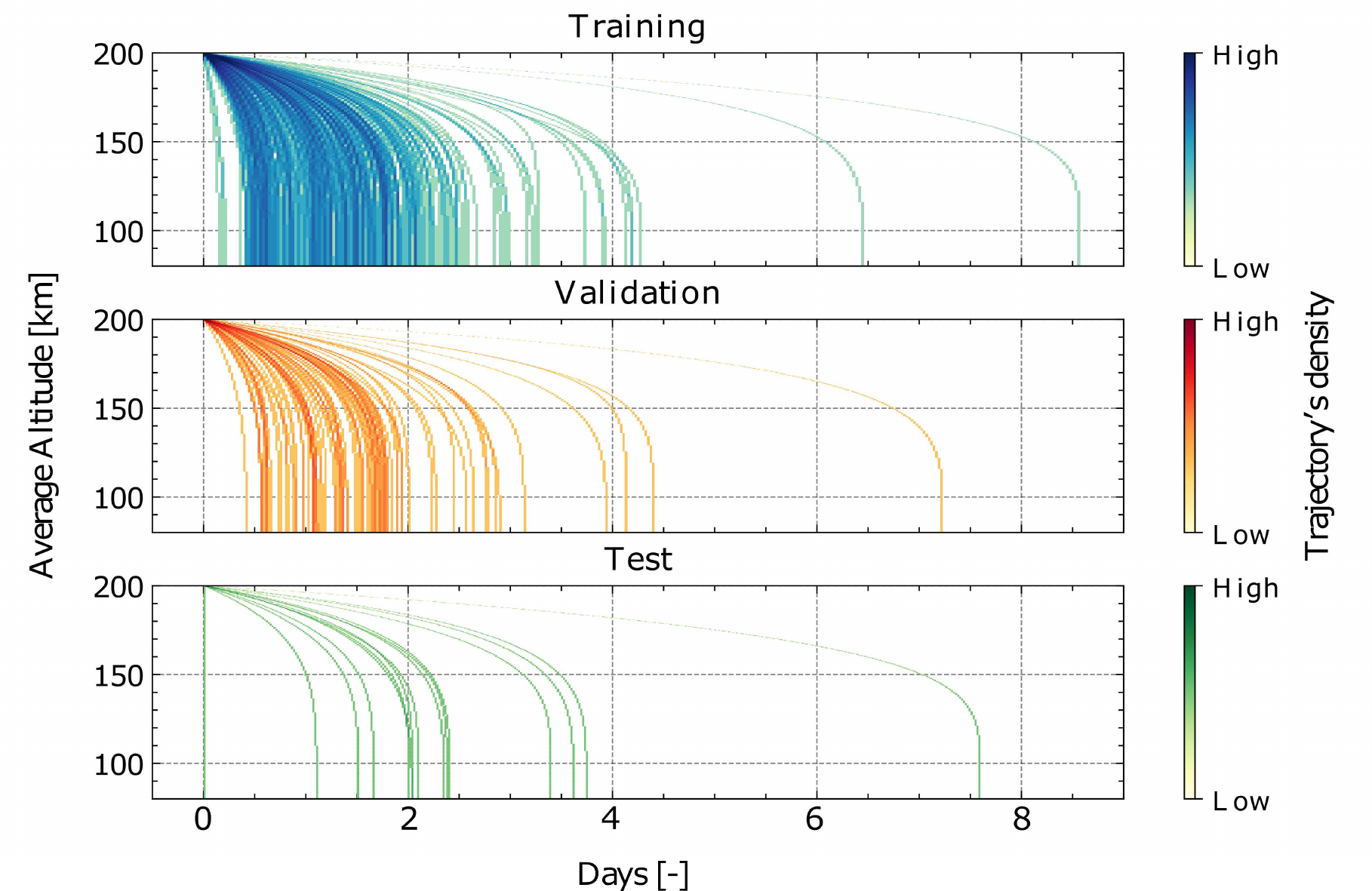}
    \caption{Trajectories generated with the fitting and sampling procedure for the training (top), validation (middle), and test (bottom) sets. Darker shadings are associated to higher concentrations of trajectories for a given re-entry time.}
    \label{fig:trajectories}
\end{figure}

The average altitude feature is of fundamental importance for deep-learning-driven re-entry prediction. In fact, it is related to the definition of the loss function used to train the RNN \cref{eq:loss_mse}. This means that it is used both as input and as
output for the deep learning model, such that given a part of the average altitude profile as input, the remaining part is predicted. The pre-processing of the average altitude profile follows the work of Jung et al. \cite{jung2021recurrent}. In particular, each trajectory is generated by fitting the available TLEs, below 240 km, according to the following equation:

\begin{align}
    \begin{split}
    f(t)&={}a_{1}+a_{2}\left(t_{r e f}-t\right)^{\frac{1}{2}}+a_{3}\left(t_{r e f}-t\right)^{\frac{1}{3}} \\
        &\quad +a_{4}\left(t_{r e f}-t\right)^{\frac{1}{4}}
    \end{split}\label{eq:fitting_us}
\end{align}

where $t_{ref}$ is the re-entry time, and $a_1$, $a_2$, $a_3$ and $a_4$, are the parameter obtained through curve fitting using the Levenberg-Marquardt algorithm, which is available through the python library SciPy \citep{2020SciPy-NMeth}. The first coefficient, $a_1$, is equal to 80 km, which is the reference altitude at which TIP messages are provided \citep{PARDINI201846}. This assumption avoids the fitting algorithm changing the re-entry epoch from the TIP message and, therefore, ensures consistency with the TLE data. Subsequently, the epoch corresponding to an altitude of 200 km is set as the initial epoch so that the time span from the starting altitude of 200 km to the re-entry at 80 km, identifies the residual time.\\
As each object will have different residual times, it is convenient to interpolate the residual time as a function of the average altitude. To do so, we subdivide the considered altitude range (between 200 km and 80 km) into a uniform grid of 25 points. Consequently, each trajectory is identified by a series of observations $\{x_i, t_i \}$, with $i=1,2,...\,25$, where $x_i$ represents the average altitude and $t_i$ the relative epoch. In this way, it is possible to use only $t_i$ as the output of the deep learning model. \cref{fig:fit_sample_traj} shows a result of such a procedure for the object identified by NORAD 27392. The resulting trajectories are represented for the training, validation, and test sets are represented in \cref{fig:trajectories}. It can be seen that there is a concentration of trajectories characterised by residual lifetimes between 0.5 days and 3 days. Meanwhile, only 3 trajectories with residual lifetimes higher than 6 days are present, and only two of them are used for training. This is important because this distribution incorporates the ballistic coefficient information and it can be directly related to the prediction capability of the model. We also note that in the test set, the leftmost trajectory has a residual time of roughly 14 minutes and it is associated with the Molniya 3-39 spacecraft. In this case, the trajectory is highly elliptical and does not satisfy our selection criterion defined in \cref{sec:data_processing}; however, it has been included in the test set to assess the capabilities of the model with such trajectories.

In a real case scenario where the entire trajectory is not known a priori, \cref{eq:fitting_us} can be used with $t_{ref}$ as the epoch of the last available TLE. However, the resulting curve will be different from the one obtained with all the available TLEs. We thus expect to observe some differences in performance between the training and testing sets. Further analysis should be carried out on this aspect because there may be variations in the predicted trajectory that depends on the adopted procedure and on the number of available TLEs. However, this difference should be minimised when a high number of TLE points is available, which is the case, for example, of the objects analysed during IADC campaigns.

\subsection{Ballistic coefficient and B*}
\label{subsec:bstar}
The ballistic coefficient, $B$, has a key importance on the re-entry of low eccentricity orbits because the dynamic is dominated by the atmospheric drag perturbation. The ballistic coefficient, as defined in \cref{eq:ballistic_coefficient_definition} \cite{curtis2013orbital}, describes the aerodynamic interaction of the spacecraft with the atmosphere via its shape, mass, and drag coefficient.

\begin{equation}
    B = C_D\frac{A}{m}
    \label{eq:ballistic_coefficient_definition}
\end{equation}

\begin{figure*}[htb!]
   \centering
   \includegraphics[width=0.7\linewidth]{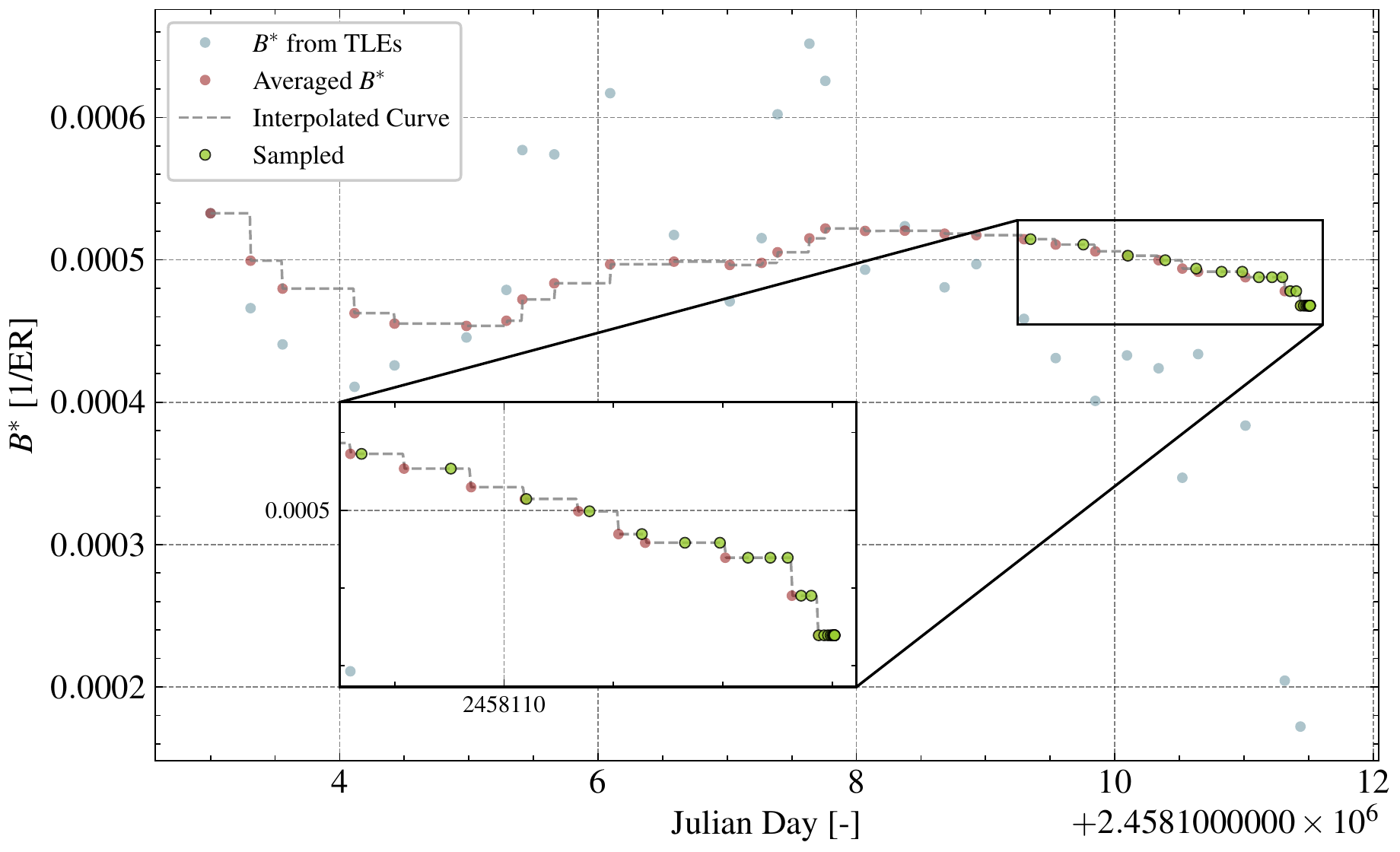}
   \caption{Representation of the pre-processing of the $B^*$ feature generation: step-wise linear interpolation with moving window average.}
   \label{fig:interpolation_bstar}
\end{figure*}

However, as we are using TLE data for the re-entry prediction, we use the drag-like coefficient, $B^*$, as a proxy of the ballistic coefficient. In fact, $B^*$ is the coefficient introduced in the SGP4 propagator, which in turn is used for the generation of TLE data. The expression for the $B^*$ is as follows \cite{vallado2001fundamentals}:

\begin{equation}  \label{eq:bstar}
B^{*}=\frac{1}{2} \frac{C_{D} A}{m} \rho_{o} R_{\oplus}
\end{equation}

where $\rho_0$ is the atmospheric density at the perigee of the orbit, assumed as $2.461\times10^{-5}\,\mathrm{kg} / \mathrm{m}^{2} / \mathrm{ER} \text{, with } \mathrm{ER=6375.135 km}$; and $R_{\oplus}$ the Earth radius of 6378.135 km.  In general, the $B^*$ coefficient of the SGP4 formulation soaks up different modelling errors. Despite the ballistic coefficient can be retrieved from $B^*$, they are not identical. However, in this context, we assume $B^*$ can be used as a proxy of $B$ as we are considering objects with similar orbital parameters that will be subject to similar modelling uncertainties. Similarly to the average altitude, it is essential to pre-process the $B^*$ data in order to feed them to the RNN. The first step of the processing consists of defining a sliding window with a fixed size to create a series of average values, in which each term is given by the average of all the previous points. The second step consists in applying a linear piece-wise interpolation scheme, in which, between two consecutive values, the preceding one is kept constant. Subsequently, the interpolated curve is sampled at the epochs defined through the average altitude processing. An example of the aforementioned procedure is represented in \cref{fig:interpolation_bstar} for NORAD 27392.

\begin{figure}[htb!]
\centering
    \includegraphics[width=0.8\columnwidth]{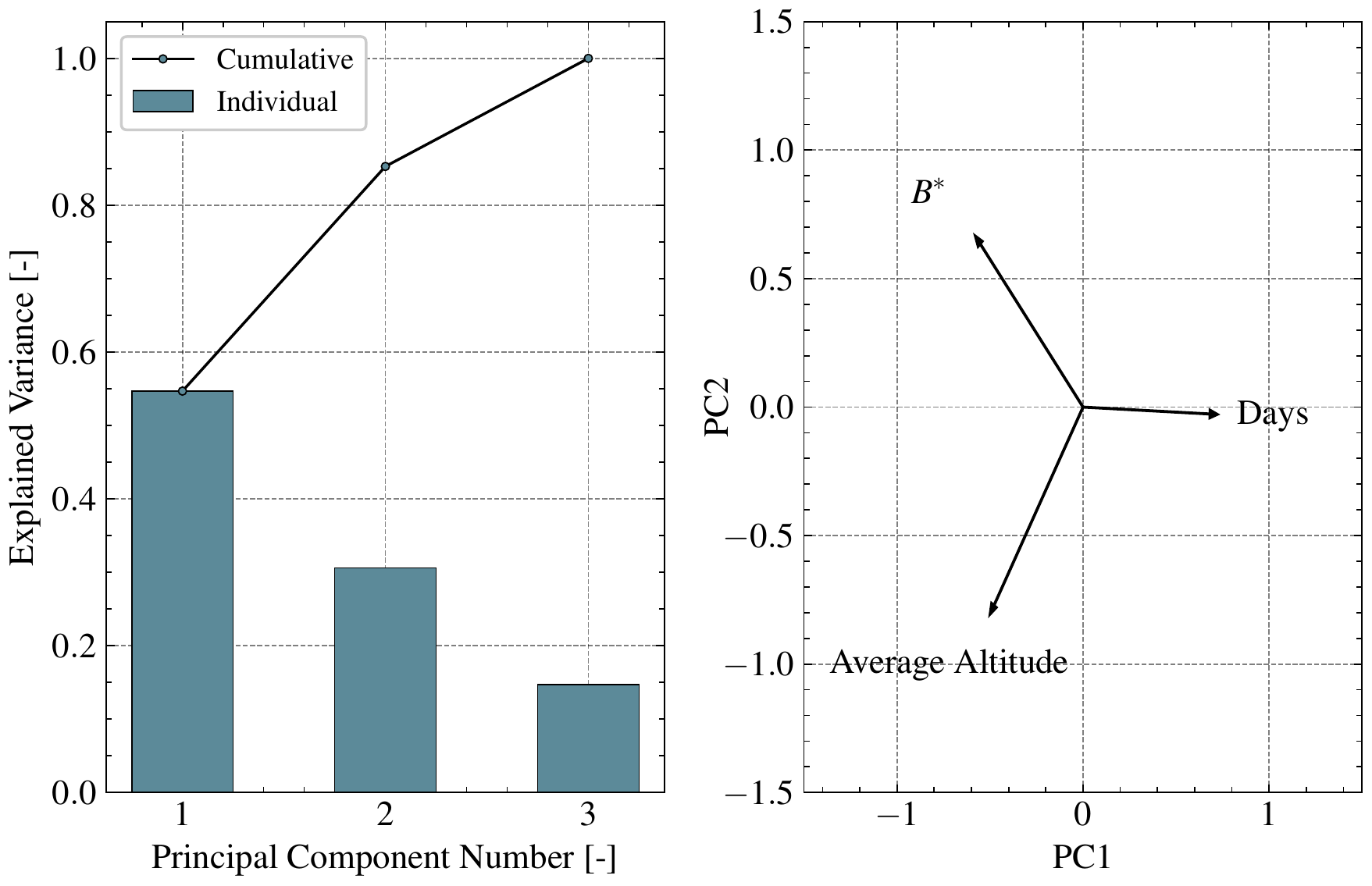}
    \caption{Variance plot (Left) and loading plot (Right) with $B^*$, residual time and altitude.}
    \label{fig:pca_1}
\end{figure}

Given the features generated so far, it is possible to apply the PCA to observe if the expected relations between the $B^*$, the residual time in days and average altitude exist. The results of the analysis are summarised in \cref{fig:pca_1}. Focusing on the loading plot and the PC1, the feature related to the time has a strong positive loading; meanwhile, the others have negative loadings. Indeed, the time feature can be seen as a synonym of the residual lifetime, therefore as the $B^*$ increases, the lifetime of an object decreases due to the drag acceleration growth. Furthermore, the altitude-time relation is associated with the altitude decrease, as time proceeds.

\subsection{Solar Index}
\label{subsec:solar_index}
\begin{figure*}[htb!]
    \centering
    \includegraphics[width=\columnwidth]{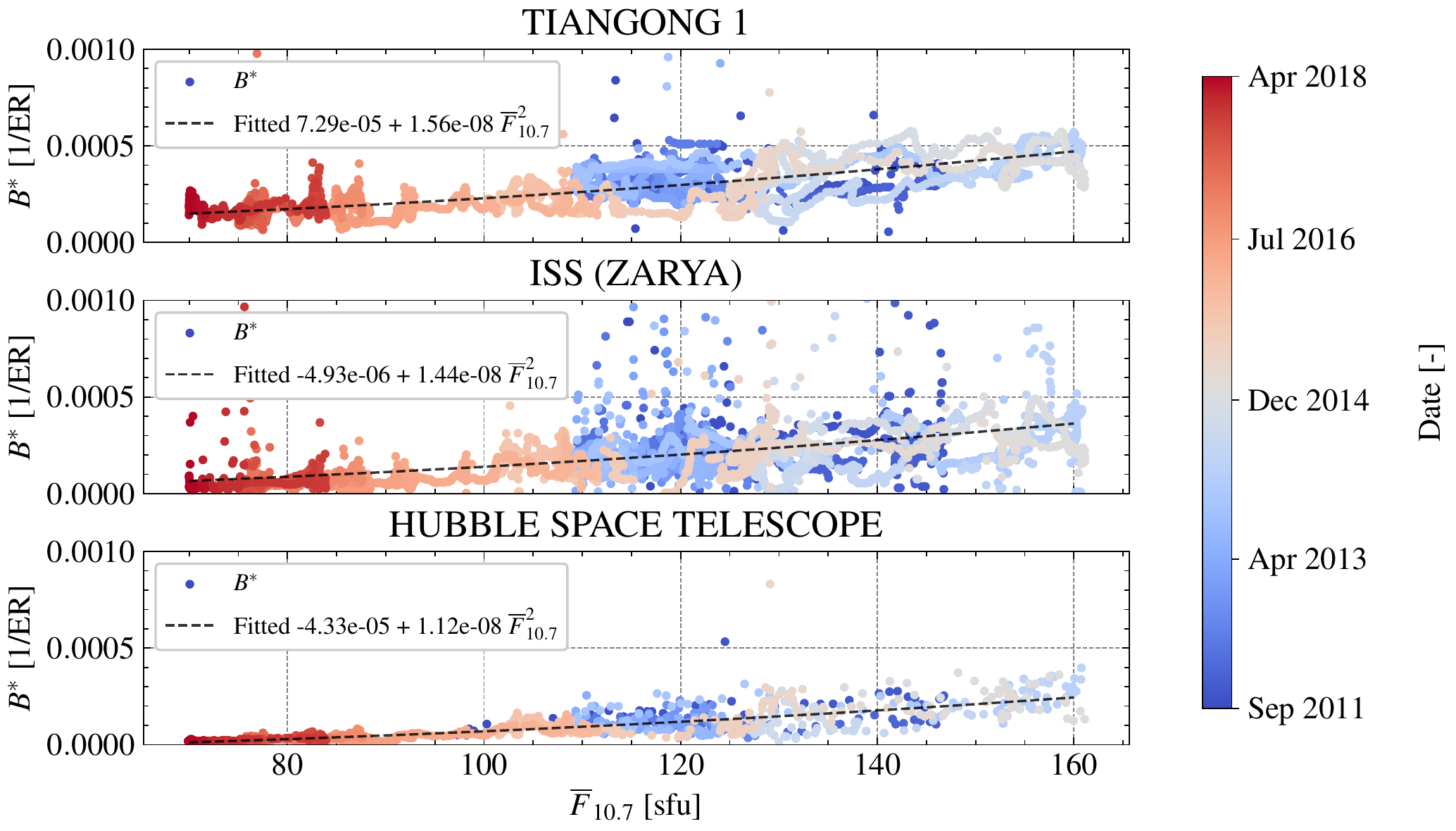}
    \caption{$B^*$ as a function of the mean solar index for Tiangong 1, ISS and HST.}
    \label{fig:bstar_as_solar}
\end{figure*}

The Solar Index, $F_{10.7}$, is a parameter that describes the intensity of solar radiation. It is important for predicting the variations of the atmospheric density \cite{vallado2001fundamentals}. The data from TLEs do not include the atmospheric density; however, the $B^*$ accounts also for the solar index variations. To highlight this dependence, \cref{fig:bstar_as_solar} represents the drag-like coefficient as a function of the last 81-day average of the solar index $\overline{F}_{10.7}$ for different bodies. It can be observed that, as $\overline{F}_{10.7}$ increases, the $B^*$ growths with a non-linear behaviour, showing the existence of a relation between the variables. In addition, this is also highlighted by the same colours of $B^*$ groups for the same values of $\overline{F}_{10.7}$, and by the terms of the curve fitting associated with $\overline{F}_{10.7}^2$. Therefore, the $\overline{F}_{10.7}$ index is introduced as a feature of the RNN, along with the area-to-mass ratio. This decision stems from the fact that the model must deal with objects with similar $B^*$ but completely different solar indexes. From an implementation point of view, it has been found that including only the $\overline{F}_{10.7}$ index corresponding to the starting epoch of the prediction led to the highest performance for the RNN. Therefore, it is included and repeated in the input sequence as a constant value. The same procedure is applied to the area-to-mass ratio, which is kept fixed.

\begin{figure}[htb!]
\centering
    \includegraphics[width=0.8\columnwidth]{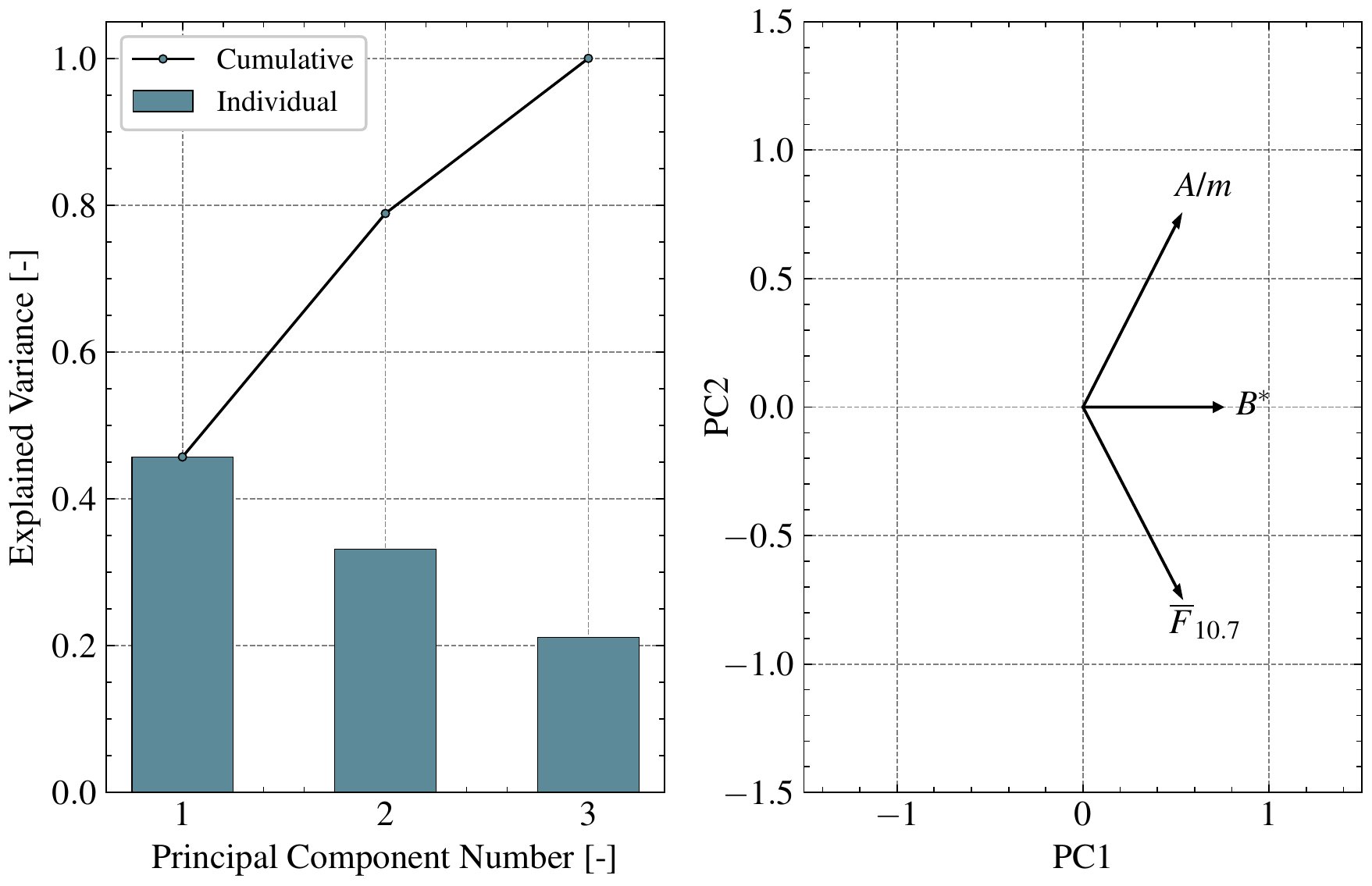}
    \caption{PCA characteristics of Solar index, $B^*$ and area-to-mass ratio.}
    \label{fig:pca_characteristics_solar}
\end{figure}

Similarly to \cref{subsec:hmean,subsec:bstar}, it is possible to explore the features generated so far in order to find if the expected relations can be found. To this aim, the PCA can be applied considering the $\overline{F}_{10.7}$ index, the $B^*$ and the area-to-mass ratio. The results are summarised in \cref{fig:pca_characteristics_solar}. From the loading plot, it can be seen that, along PC1, all the variables are characterised by positive loadings. In particular, the relation between the $A/m$ and the $B^*$ is related to the drag-like coefficient definition. The relation between the $B^*$ and the solar index is the same as discussed in \cref{fig:bstar_as_solar}, where a positive correlation is present. Moreover, the relation between the area-to-mass ratio and the $\overline{F}_{10.7}$ can be explained through the $B^*$. 

\subsection{Data regularisation}
The features generated so far include the average altitude profile, $B^*$, the $\overline{F}_{10.7}$ and the area-to-mass ratio. Due to their nature, the solar index and time-related feature are normalised according to \cite{raschka2015python}:

\begin{equation}
    \bar{\bm{x}} = \frac{\bm{x} - \min(\bm{x})}{\max(\bm{x}) - \min(\bm{x})}
\end{equation}

where $\bar{\bm{x}}$ represents the normalised feature; and $\bm{x}$ the related original input feature.

\section{Results}
\label{sec:results}

The Seq2Seq architecture described in \cref{sec:deep_model} is independently trained on four different cases, characterised by different starting altitudes, as shown in \cref{tab:characteristics_training}. In this way, the sensitivity of the prediction accuracy with respect to the starting altitude can be assessed.

\begin{table}[htb!]
\centering
\begin{tabular}{c p{2.2cm} p{2.3cm} p{2.2cm}}
\hline 
\hline
\textbf{Case}       & \boldsymbol{$T_x$} [-] & \textbf{Starting altitude} [$km$] & \textbf{Training epochs} [-]\\
\hline
A    &  5   & 180 & 2900\\
\hline
B &  9    & 160 & 3000\\
\hline
C &  13   &  140 & 1200\\
\hline
D &  17   &   120 & 1200\\
\hline 
\hline
\end{tabular}
\caption{Characteristics of training cases.}
\label{tab:characteristics_training}
\end{table}

In each one of the cases in \cref{tab:characteristics_training}, the entire set of TLEs is split into two independent parts, the training and validation sets. Furthermore, both the inputs and outputs of the two sets have the same lengths. Therefore, the model is trained to predict the output sequence of length $T_y$, given an input sequence of length $T_x$, and the same strategy is also applied for validation. This means that the losses, $\mathcal{L}_{MSE}^{Val}$ and $\mathcal{L}_{MSE}^{Train}$, are computed on the values associated with the same average altitudes. The only difference is represented by the objects analysed, which are different between the sets. The input dataset is mathematically represented by a tensor of rank 3, which, in TensorFlow's notation, has dimension [$\text{Objects} \times \text{Time} \times \text{Features}$]. Therefore, according to the problem definition, the model is faced only with a part of the input trajectory of the objects, together with all the related input features. The length of the time dimension can be chosen according to the desired needs, by varying the value of $T_x$, which is associated with the starting altitude of the prediction. Hence, the five different cases in \cref{tab:characteristics_training} are identified by different lengths of the input sequence $T_x$. Regarding the output, the model has to predict the final part of the average trajectory altitude until the re-entry epoch. The sum of the input and output sequence is $T_x + T_y = 25$.

\subsection{Hyperparameters optimisation}
The deep learning model described in \cref{sec:deep_model} requires the definition of specific hyperparameters that are variables on which the learning algorithm relies for its optimisation and prediction processes. In general, hyperparameters may be different depending on the problem in exam; therefore, it is necessary to perform an informed selection of such parameters. This is usually achieved by optimising the hyperparameters in order to maximise the performances of the deep learning model. However, the hyperparameters optimisation process can be computationally intensive; therefore, for this work, it has been decided to perform the optimisation only of Case A of \cref{tab:characteristics_training}. In fact, this case is characterised by the smallest input sequence and is therefore considered the most challenging of the four cases. The optimisation is designed using \textit{HyperOpt} \citep{pmlr-v28-bergstra13} module, that is Bayesian optimisation library, together with the Asynchronous Successive Halving (ASHA) algorithm \citep{li2020massively} to combine random search with the early stopping of hyperparameters combinations that exhibit poor performances. These algorithms can be conveniently combined and parallelised using the Python package \textit{Ray Tune} \citep{liaw2018tune}.
In this work, part of the hyperparameter space has been kept constant and another part has been optimised. \cref{tab:hyper_model_fixed} shows the fixed hyperparameters and their associated value. As shown in \cref{eq:loss_mse}, the loss function is the Mean Squared Error; the selected optimiser is Adam \citep{kingma2014adam}, which is a commonly used algorithm for deep learning model weights optimisation. It is a first-order gradient-based optimisation, based on adaptive estimates of first and second-order moments (i.e., mean and variance). As specified in \cref{tab:hyper_model_fixed}, the decay rate of the first order moment, $\beta_1$, and of the second order moment, $\beta_2$, have been set, together with the \textit{Clipnorm} parameters, which forces a re-scaling of the gradient whenever its norm exceeds the specified value.

\begin{table}[htb!]
\centering
\begin{tabular}{lll}
\hline \hline
                           & \textbf{Parameter}       & \textbf{Value [-]} \\ \midrule
\multirow{5}{*}{\textbf{Fixed}}     & Loss                     & Mean Squared Error \\ \cmidrule(l){2-3} 
                           & Optimiser                & Adam               \\ \cmidrule(l){2-3} 
                           & $\beta_1$                & 0.999              \\ \cmidrule(l){2-3} 
                           & $\beta_2$                & 0.999              \\ \cmidrule(l){2-3} 
                           & Clipnorm                 & 0.1                \\
\hline \hline
\end{tabular}
\caption{Fixed hyperparameters.}
\label{tab:hyper_model_fixed}
\end{table}

The remaining hyperparameters that are optimised are summarised in \cref{tab:search_space}, together with the relevant search spaces and sampling mode. The parameter related to the layer is defined as the number of stacked GRU units of the encoder and decoder, respectively. This means that if the number of layers is 2, the model is going to be composed of 4 units, where 2 are associated with the encoder and the remaining with the decoder. Specifically, the final hidden state of each layer of the encoder is passed to their respective layer of the decoder. For example, the first GRU unit of the decoder is initialised with the hidden state of the first unit of the encoder.

\begin{table}[!htb]
\centering
\begin{tabular}{c c c c}
\hline 
\hline
\textbf{Hyperparameter}       & \textbf{Interval} [min, max] &  \textbf{Sampling Mode}\footnotemark\\
\hline
Learning Rate    &  [$1\cdot10^{-6}$, $1\cdot10^{-1}$]  & Uniform Logarithmic \\
\hline
Number of layers &  [1, 3]  & QUniform \\
\hline
Hidden size GRU &  [16, 256] & QUniform \\
\hline 
Batch Size & [8, 64] & QUniform \\
\hline
Decay Scheduled Sampling & [$1\cdot10^{-1}$, $9.9\cdot10^{-1}$] & Uniform Logarithmic\\
\hline
\hline
\end{tabular}
\caption{Optimised hyperparameters and their relevant search space in \textit{HyperOpt}.}
\label{tab:search_space}
\end{table}
\footnotetext{It refers to the particular distribution from which the hyperparameter is selected. See \url{http://hyperopt.github.io/hyperopt/getting-started/search_spaces/} for further details.}

Finally, it is also necessary to specify the parameters for the ASHA algorithm. \cref{tab:asha_input} shows the aforementioned parameters and the selected values. Specifically, the total number of trials refers to the amount of hyperparameters combinations the algorithm attempts before stopping, the reduction factor represents the fraction of trials removed after each run and the grace period that corresponds to the epochs for which a trial must be trained before halving. These values have to be selected considering the computational resources of the hardware and the expected results. The most suitable parameters, which have been found with a trade-off between results and computational time, are represented in \cref{tab:asha_input}. It has to be noted that a small grace period may bias ASHA towards those solutions that converge quite rapidly but do not lead to the best performances. Therefore, a good compromise is experimentally found to be at 400 epochs. Furthermore, the maximum value of the training epoch is set at 2100 because it has been observed that the optimal value is typically above 2000.

\begin{table}[!htb]
\centering
\begin{tabular}{c c}
\hline 
\hline
\textbf{Input}       & \textbf{Value} \\
\hline
Number of Trials    &  100 [-]\\
\hline
Reduction Factor &  4 [-]\\
\hline
Grace Period & 400 [Epochs]\\
\hline
Maximum Training Epoch & 2100 [Epochs]\\
\hline
\hline
\end{tabular}
\caption{Input parameters of ASHA.}
\label{tab:asha_input}
\end{table}

\begin{figure}[!htb]
    \centering
    \includegraphics[width=0.8\columnwidth]{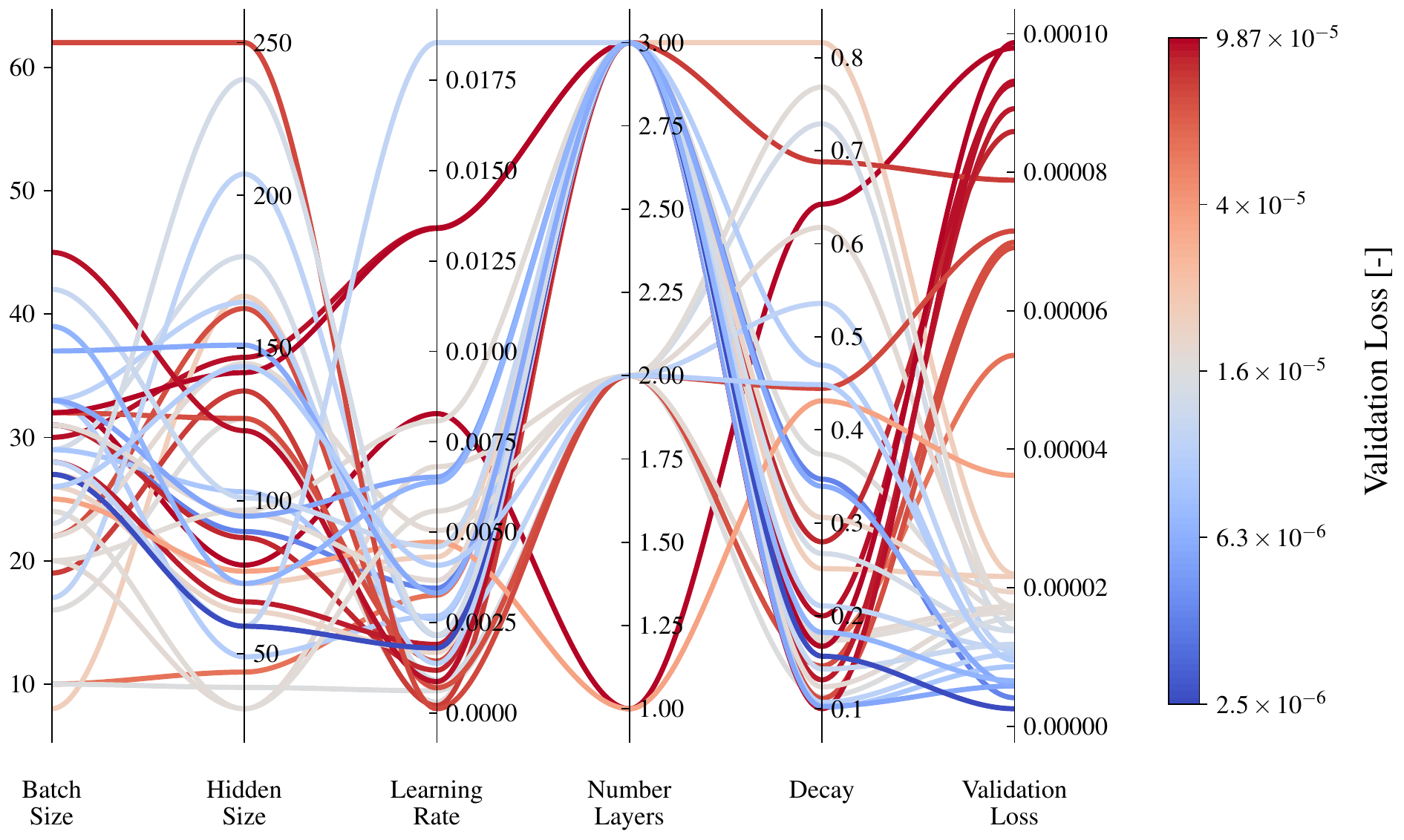}
    \caption{Representation of selected hyperparameter combinations as parallel lines that pass to the vertices corresponding to the hyperparameters associated to each trial, where each colour is related to the validation loss.}
    \label{fig:paral_coor}
\end{figure}

\cref{fig:paral_coor} shows the validation losses as a function of the value of each hyperparameter. We can observe that the curves associated with lower losses are approximately distributed at low values of the batch size, between 20 and 40, and hidden state size, in particular between 50 and 150. Furthermore, these distributions provide the qualitative importance of each parameter. Indeed, an important parameter can be defined by observing the magnitude changes of the resulting loss with respect to a variation of its value. This means that a change, even small, in such an input parameter, determines an important variation of the resulting loss. For example, by looking at the learning rate in the parallel coordinates plot, it can be seen the distinction between red and blue curves, where the lower rates lead to high validation losses. Instead, values between 0.002 and 0.0075 seem to lead to better performances. Contrarily, the decay rate of the Scheduled Sampling mechanism appears to have a negligible effect on the validation loss. Indeed, it is not possible to identify a range of values that leads to higher performances. This is related to the fact that the decay rate in \cref{eq:decay_SS} does not have a significant importance on the convergence because the threshold, for selecting the ground truth or the model output, decreases quickly for all the values that the optimiser selected, forcing the model to learn from its predicted outputs. Therefore, it may suggest that the model performs better when it has to deal with its own errors, in a way that is more similar to inference than training. 

Finally, the optimised values of the hyperparameters are given in \cref{tab:hyper_results}.

\begin{table}[htb!]
\centering
\begin{tabular}{lll}
\hline \hline
                           & \textbf{Parameter}       & \textbf{Value [-]} \\ \midrule
\multirow{5}{*}{\textbf{Optimised}} & Learning Rate            & 0.001795           \\ \cmidrule(l){2-3} 
                           & Number of Layers         & 3                  \\ \cmidrule(l){2-3} 
                           & Hidden Size GRU          & 59                 \\ \cmidrule(l){2-3} 
                           & Batch Size               & 27                 \\ \cmidrule(l){2-3} 
                           & Decay Scheduled Sampling & 0.15665            \\ \hline \hline
\end{tabular}
\caption{Optimised hyperparameters.}
\label{tab:hyper_results}
\end{table}

\subsection{Dataset characterisation}
\label{subsec:dataset_characterisation}

\begin{figure*}[!htb]
    \centering
    \includegraphics[width=\columnwidth]{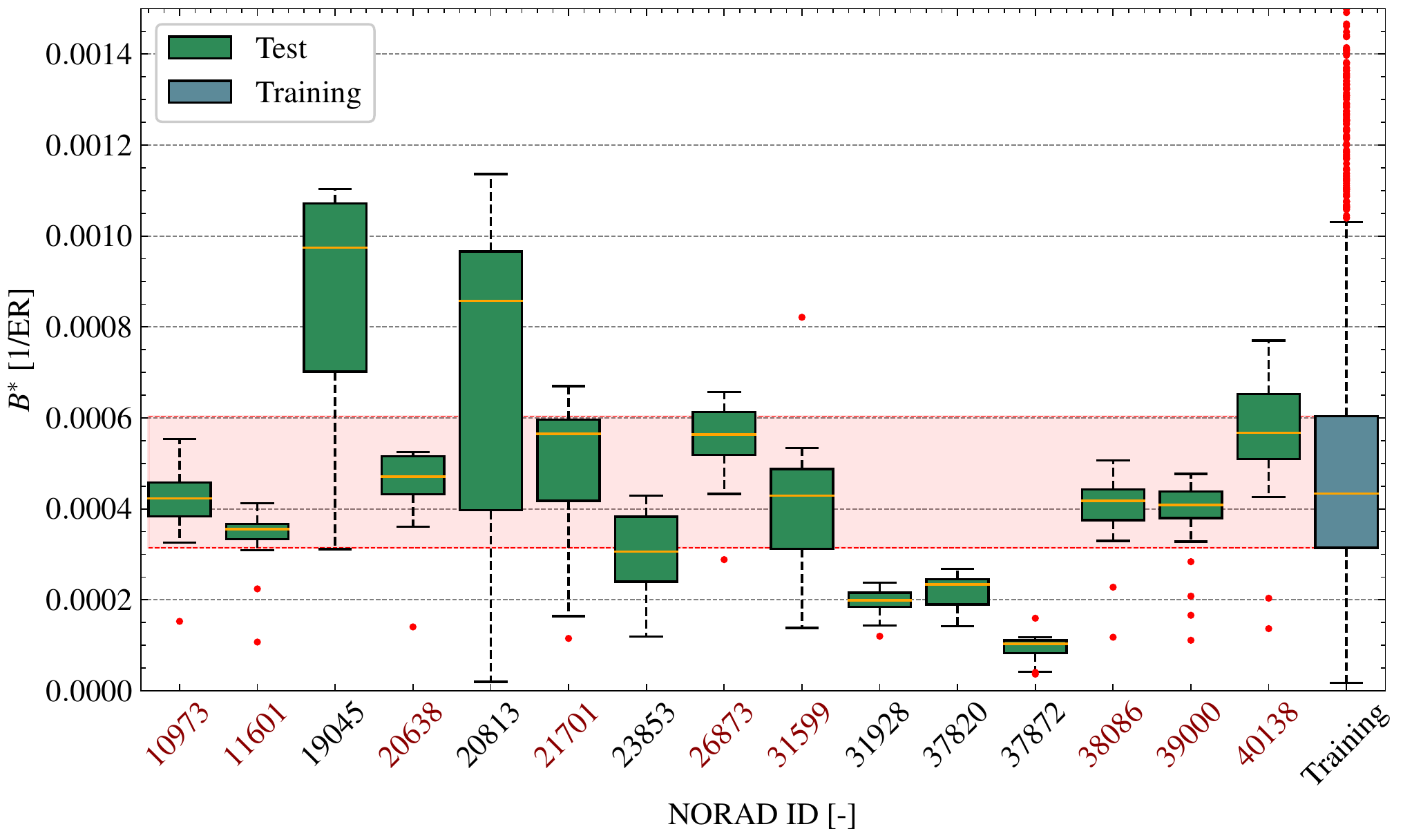}
    \caption{$B^*$ distribution of objects in training and test set, where red NORAD IDs identify test bodies with similar $B^*$ distributions as the training set (Category 1).}
    \label{fig:boxplot_test_training}
\end{figure*}

To analyse and understand the performances of the model, the test objects have to be divided depending on their physical properties with respect to the training set. We define two categories of objects as in \cref{tab:characteristics_training} (\cref{subsubsec:test_obj_categories}) in which the objects of the test set share similar characteristics to the training set. Specifically, as shown in \cref{fig:boxplot_test_training}, the two categories are subdivided with respect to the $B^{*}$ coefficient: \textit{Category 1} objects (red label) have a median $B^{*}$ coefficient inside the 0.25 and 0.75 quantiles of the training set distribution, while the median $B^{*}$ of objects in \textit{Category 2} (black label) lies outside this range. Therefore, given the significant importance of $B^*$, it is expected that the deep learning model is going to perform better with those objects for which their distributions are similar to the training.

\begin{figure*}[!htb]
    \centering
    \includegraphics[width=\columnwidth]{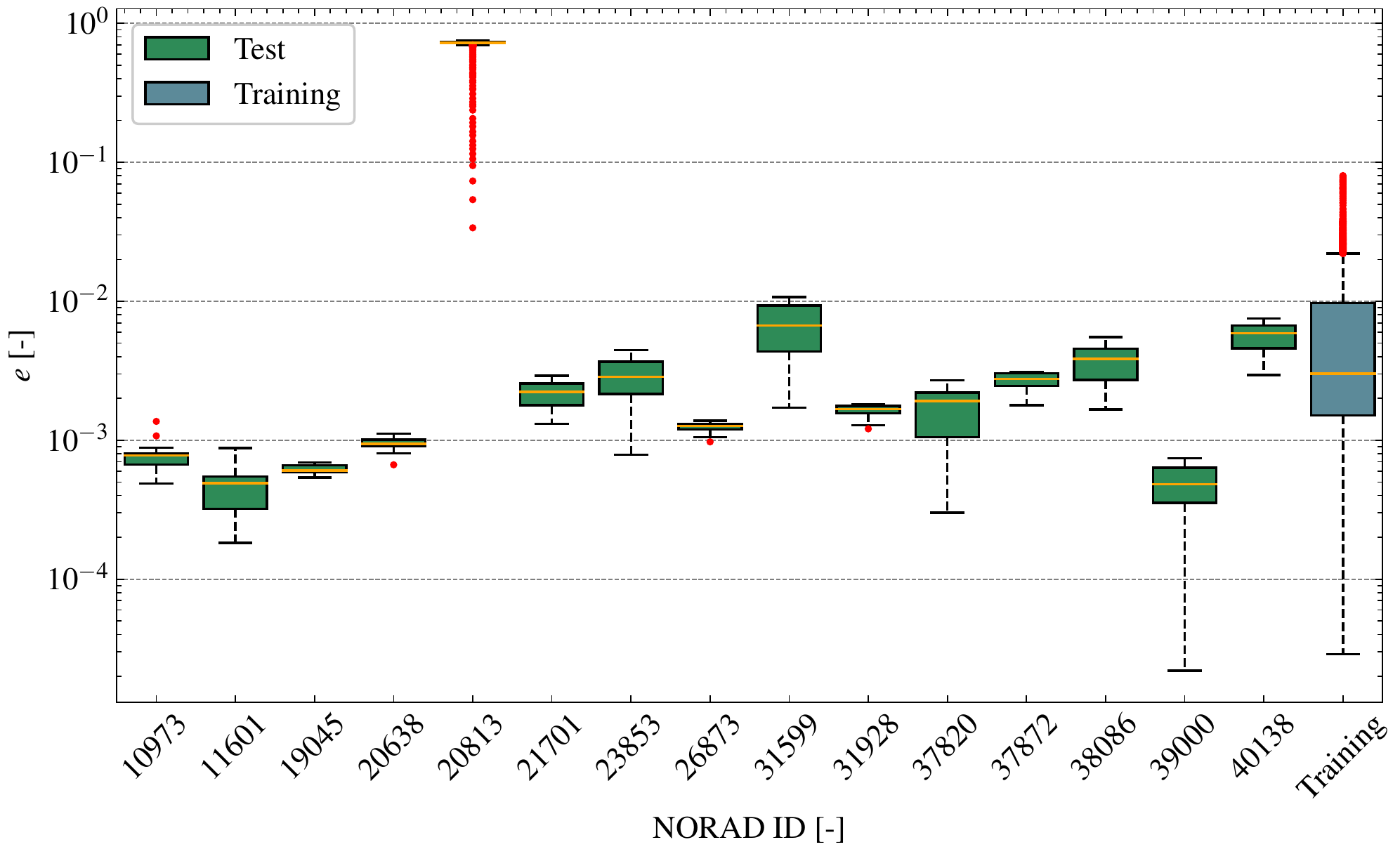}
    \caption{Eccentricity distribution of objects in training and test set.}
    \label{fig:box_plot_ecc}
\end{figure*}

To analyse and understand the performances of the model, the test objects have to be divided depending on their physical properties with respect to the training set. Hence, \cref{fig:boxplot_test_training} represents the $B^*$ distributions of the objects in the test and training sets, where two categories of bodies can be identified, as highlighted by the different colours of the NORAD IDs. Another parameter to analyse is the eccentricity, which distributions for the training and test set are represented in \cref{fig:box_plot_ecc}. It can be seen that 20813 represents an anomaly with respect to all the other objects. Indeed, it is characterised by a median of roughly 0.7, which identifies it as a highly elliptical orbit, for which the nature of the re-entry is different. However, we have decided to maintain this object to further test the capabilities and limitations of the developed deep learning model.

\bigbreak
To characterise the performances and compare the results, let the absolute error, $\epsilon_{abs}$, and the relative error, $\epsilon_{rel}$, be respectively:

\begin{align}
    \epsilon_{abs} &= \lvert t_{Predicted} - t_{Actual} \rvert \\
    \epsilon_{rel} &= \frac{\lvert t_{Predicted} - t_{Actual} \rvert }{t_{Actual} - t_{Initial}} \times 100 
\end{align}
where $t_{Predicted}$ is the predicted re-entry epoch; and $t_{Actual}$ is the assessed re-entry epoch.

\subsection{Testing}
\label{subsec:testing}

Starting from the objects of the first category, \cref{tab:results_cat1} represents the predictions' errors for all the different starting altitudes defined in \cref{tab:characteristics_training}. Generally, it can be seen that the model is successfully capable of predicting the output trajectory and therefore the re-entry epoch. For example, 39000 shows an absolute error that is less than 2 minutes, roughly 24 hours before the re-entry. Furthermore, this error stays confined below roughly 3 minutes, for all the predictions. This behaviour is related to the similarities of the $B^*$ distributions of each object with respect to the one of the training set. Indeed, it should be highlighted that the highest absolute error is 14 minutes and it is associated with 40138. Moreover, observing the MSEs, all the relative errors lay between $10^{-5}$ and $10^{-7}$, showing that the similarities of the $B^*$ are associated with excellent performances. It is interesting to note that comparing the different cases, the MSE decrease does not necessarily correspond to a lower absolute error, as for example in 11601 and 39000. This indicates that better results could also be obtained by selecting a different loss function.

\begin{table*}[!htb]
\centering
\begin{adjustbox}{width=\linewidth,center}
\begin{tabular}{@{}llccccccccc@{}}
\hline \hline
\textbf{Case}                               & \multicolumn{1}{c}{\textbf{Metric}} & \multicolumn{9}{c}{\textbf{NORAD}}                                                                        \\ \midrule
                                            &                                     & \textbf{10973}     & \textbf{11601}     & \textbf{20638}     & \textbf{21701}     & \textbf{26873}     & \textbf{31599}     & \textbf{38086}     & \textbf{39000}     & \textbf{40138}     \\ \midrule
\multicolumn{1}{c}{\multirow{3}{*}{Case A}} & \textbf{$\boldsymbol{\varepsilon_{abs}}$ [hrs]}  & 0.0129    & 0.0012    & 0.0593    & 0.0287    & 0.0381    & 0.0835    & 0.0856    & 0.0263    & 0.2225    \\ \cmidrule(l){2-11} 
\multicolumn{1}{c}{}                        & \textbf{$\boldsymbol{\varepsilon_{rel}}$ [\%]}   & 0.0582    & 0.0051    & 0.2930    & 0.1262    & 0.1800    & 0.5004    & 0.3469    & 0.1093    & 1.2464    \\ \cmidrule(l){2-11} 
\multicolumn{1}{c}{}                        & \textbf{MSE [$\text{day}^2$]}       & 4.4482e-6 & 1.1707e-6 & 1.0218e-5 & 1.0236e-5 & 1.7067e-6 & 5.2804e-6 & 2.1809e-5 & 1.0643e-6 & 1.9246e-5 \\ \midrule
\multirow{3}{*}{Case B}                     & \textbf{$\boldsymbol{\varepsilon_{abs}}$ [hrs]}  & 0.0915    & 0.0411    & 0.0417    & 0.0021    & 0.0376    & 0.0640    & 0.0187    & 0.0191    & 0.0191    \\ \cmidrule(l){2-11} 
                                            & \textbf{$\boldsymbol{\varepsilon_{rel}}$ [\%]}   & 0.9903    & 0.4878    & 0.6032    & 0.0265    & 0.4828    & 1.0072    & 0.2198    & 0.2134    & 0.2952    \\ \cmidrule(l){2-11} 
                                            & \textbf{MSE [$\text{day}^2$]}       & 7.0168e-6 & 1.8629e-6 & 2.6473e-6 & 4.4563e-6 & 2.6635e-7 & 1.2836e-6 & 1.1987e-6 & 1.7427e-6 & 1.8259e-6 \\ \midrule
\multirow{3}{*}{Case C}                     & \textbf{$\boldsymbol{\varepsilon_{abs}}$ [hrs]}  & 0.0934    & 0.0043    & 0.0125    & 0.0268    & 0.0610    & 0.1025    & 0.0368    & 0.0065    & 0.0198    \\ \cmidrule(l){2-11} 
                                            & \textbf{$\boldsymbol{\varepsilon_{rel}}$ [\%]}   & 2.8694    & 0.1839    & 0.7061    & 1.3862    & 2.8083    & 5.6717    & 1.7670    & 0.2461    & 1.1732    \\ \cmidrule(l){2-11} 
                                            & \textbf{MSE [$\text{day}^2$]}       & 1.8350e-5 & 2.1482e-6 & 9.5146e-7 & 4.1151e-6 & 1.5597e-6 & 5.9912e-6 & 4.7233e-6 & 1.5328e-6 & 6.7500e-7 \\ \midrule
\multirow{3}{*}{Case D}                     & \textbf{$\boldsymbol{\varepsilon_{abs}}$ [hrs]}  & 0.0670    & 0.0560    & 0.0244    & 0.0340    & 0.0581    & 0.0271    & 0.0580    & 0.0420    & 0.0093    \\ \cmidrule(l){2-11} 
                                            & \textbf{$\boldsymbol{\varepsilon_{rel}}$ [\%]}   & 8.1350    & 14.1151   & 9.1041    & 13.5330   & 15.8149   & 8.8874    & 20.3210   & 8.3022    & 3.7092    \\ \cmidrule(l){2-11} 
                                            & \textbf{MSE [$\text{day}^2$]}       & 6.4009e-6 & 3.1455e-6 & 1.3601e-6 & 1.3073e-6 & 2.2189e-6 & 1.0388e-6 & 2.2950e-6 & 2.1382e-6 & 1.5021e-6 \\ \hline \hline
\end{tabular}
\end{adjustbox}
\caption{Results of objects in category 1.}
\label{tab:results_cat1}
\end{table*}

Regarding the objects in the second category, the errors for all the bodies are listed in \cref{tab:results_ca2}. It can be generally seen that the performances of the model are inferior with respect to the previous category of objects, due to the low number of bodies characterised by $B^*$ distributions similar to the one of the training set. This can be clearly observed in 37872, which can not be associated with any $B^*$ distributions in the sets. Hence, the resulting outputs are characterised by the highest MSEs, due to the fact that the model tends to overestimate all the relative epochs. However, it can be demonstrated that the output follows the shape of trajectory altitude quite accurately. Furthermore, it can be noted that 20813 is characterised by the highest relative errors and equally significant MSEs. Hence, this indicates that the model is not capable of predicting the output trajectory and it completely misses all the predicted epochs. Indeed, by inspecting \cref{fig:boxplot_test_training}, 20813 is characterised by a large distribution of $B^*$ values, which maximum is similar to 19045. However, the model has shown to follow the output profile reasonably well for 19045 in all the cases, with a maximum relative error of roughly 18 minutes and MSE between $10^{-5}$ and $10^{-7}$ [$days^2$]. However, object 20813 also has a large value of the eccentricity (see \cref{fig:box_plot_ecc}, which justifies the difficulties of the deep learning model in predicting its re-entry and indicates that, as expected, the contribution of the eccentricity cannot be ignored for orbits that are not quasi-circular.

\begin{table*}[!htb]
\centering
\begin{adjustbox}{width=0.9\linewidth,center}
\begin{tabular}{@{}llcccccc@{}}
\hline \hline
\textbf{Case}                               & \multicolumn{1}{c}{\textbf{Metric}} & \multicolumn{6}{c}{\textbf{NORAD}}                                                                  \\ \midrule
                                            &                                     & \textbf{19045} & \textbf{20813} & \textbf{23853} & \textbf{31928} & \textbf{37820} & \textbf{37872} \\ \midrule
\multicolumn{1}{c}{\multirow{3}{*}{Case A}} & \textbf{$\boldsymbol{\varepsilon_{abs}}$ [hrs]}  & 0.2994         & 1.2758         & 0.3083         & 0.6206         & 0.5829         & 1.3427         \\ \cmidrule(l){2-8} 
\multicolumn{1}{c}{}                        & \textbf{$\boldsymbol{\varepsilon_{rel}}$ [\%]}   & 2.5476         & 964.0562       & 0.8570         & 1.7715         & 1.6664         & 1.6942         \\ \cmidrule(l){2-8} 
\multicolumn{1}{c}{}                        & \textbf{MSE [$\text{day}^2$]}       & 6.1088e-5      & 1.1331e-3      & 9.9711e-5      & 3.6330e-4      & 3.2982e-4      & 1.0313e-2      \\ \midrule
\multirow{3}{*}{Case B}                     & \textbf{$\boldsymbol{\varepsilon_{abs}}$ [hrs]}  & 0.0355         & 0.7159         & 0.0273         & 0.0854         & 0.0894         & 2.5015         \\ \cmidrule(l){2-8} 
                                            & \textbf{$\boldsymbol{\varepsilon_{rel}}$ [\%]}   & 0.8068         & 1309.4202      & 0.2152         & 0.7246         & 0.7297         & 9.3210         \\ \cmidrule(l){2-8} 
                                            & \textbf{MSE [$\text{day}^2$]}       & 1.7582e-6      & 2.6453e-4      & 3.2465e-6      & 7.4105e-6      & 5.7928e-6      & 1.2797e-2      \\ \midrule
\multirow{3}{*}{Case C}                     & \textbf{$\boldsymbol{\varepsilon_{abs}}$ [hrs]}  & 0.0029         & 0.2194         & 0.0033         & 0.0080         & 0.0222         & 1.9536         \\ \cmidrule(l){2-8} 
                                            & \textbf{$\boldsymbol{\varepsilon_{rel}}$ [\%]}   & 0.2301         & 2254.3293      & 0.1022         & 0.2579         & 0.6580         & 31.1668        \\ \cmidrule(l){2-8} 
                                            & \textbf{MSE [$\text{day}^2$]}       & 1.1096e-7      & 6.2342e-5      & 6.7083e-6      & 1.6303e-5      & 1.5640e-5      & 1.1795e-2      \\ \midrule
\multirow{3}{*}{Case D}                     & \textbf{$\boldsymbol{\varepsilon_{abs}}$ [hrs]}  & 0.0805         & 0.0098         & 0.1186         & 0.0847         & 0.0788         & 2.9645         \\ \cmidrule(l){2-8} 
                                            & \textbf{$\boldsymbol{\varepsilon_{rel}}$ [\%]}   & 35.9520        & 2532.4688      & 26.520         & 16.8331        & 13.4101        & 373.4046       \\ \cmidrule(l){2-8} 
                                            & \textbf{MSE [$\text{day}^2$]}       & 9.3967e-6      & 2.5700e-5      & 2.4415e-5      & 2.3780e-5      & 1.9272e-5      & 1.5264e-2      \\ \hline \hline
\end{tabular}
\end{adjustbox}
\caption{Results of objects in category 2.}
\label{tab:results_ca2}
\end{table*}

\subsubsection{Orbit prediction - Case A}
\label{subsubsec:orbit_prediction_A}
\begin{figure}[!htb]
    \centering
    \includegraphics[width=0.8\columnwidth]{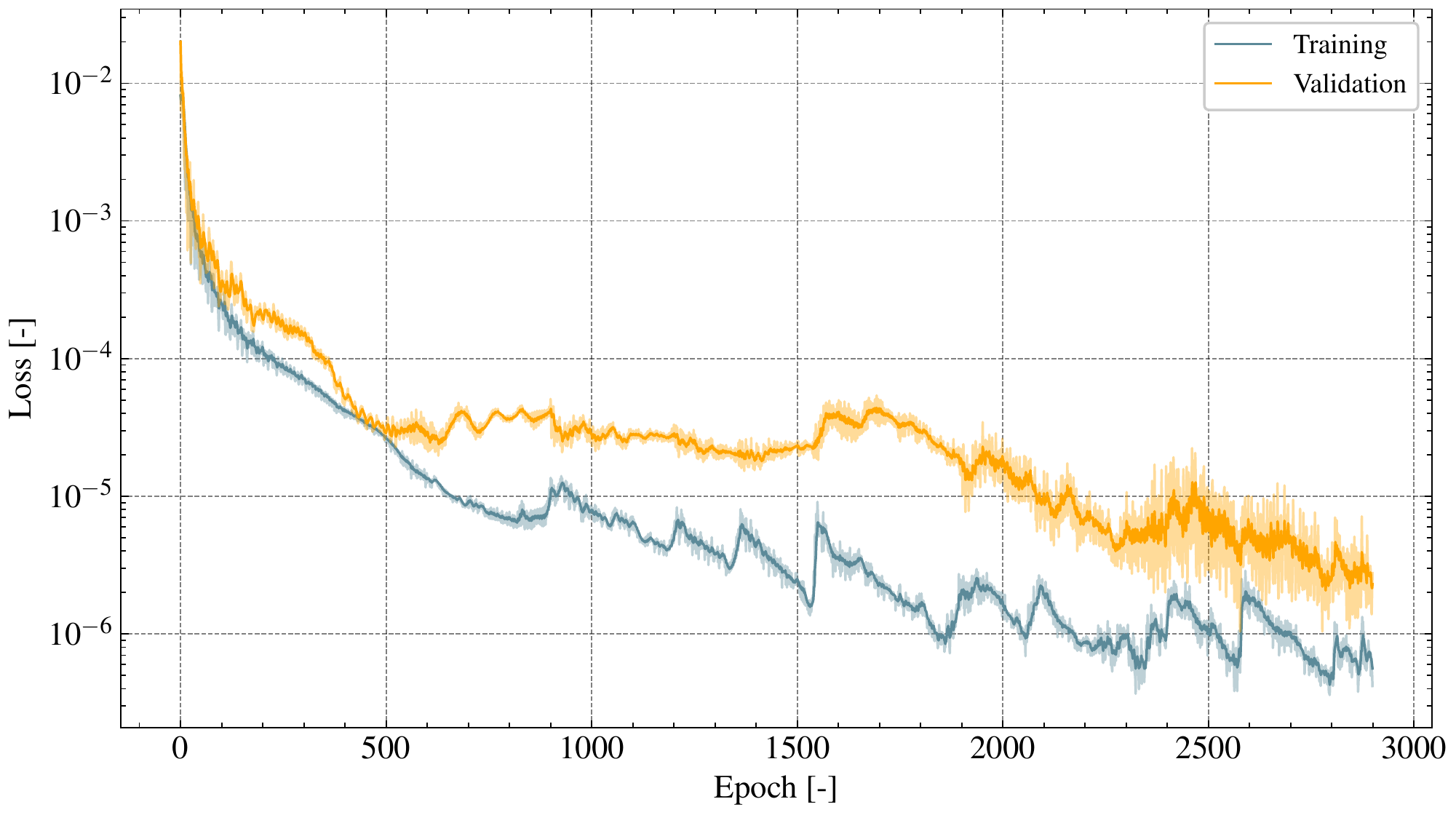}
    \caption{Representation of the training and validation loss curves in case A, where each curve is represented through its exponentially weighted average (coloured) and its original values (shaded).}
    \label{fig:training_A}
\end{figure}

In this section, we show in detail the results for Case A. Specifically, we present the evolution of the training and validation losses and the results of the trajectory predictions of the developed deep learning model for selected objects in the test set. For the test case, both categories of objects (\cref{subsubsec:test_obj_categories}) have been included. \cref{fig:training_A} shows the training and validation losses as a function of the training epoch, for a total of 2900 epochs. We observe how the rate of convergence of the validation loss is quite lower than the training loss; nonetheless, the final loss is below \num{1e-5}, which has been observed to produce accurate results.

\begin{figure}[htb!]
	\centering
	\begin{subfigure}{0.48\textwidth}
		\centering
		\label{fig:10973_A}
		\includegraphics[width=\textwidth]{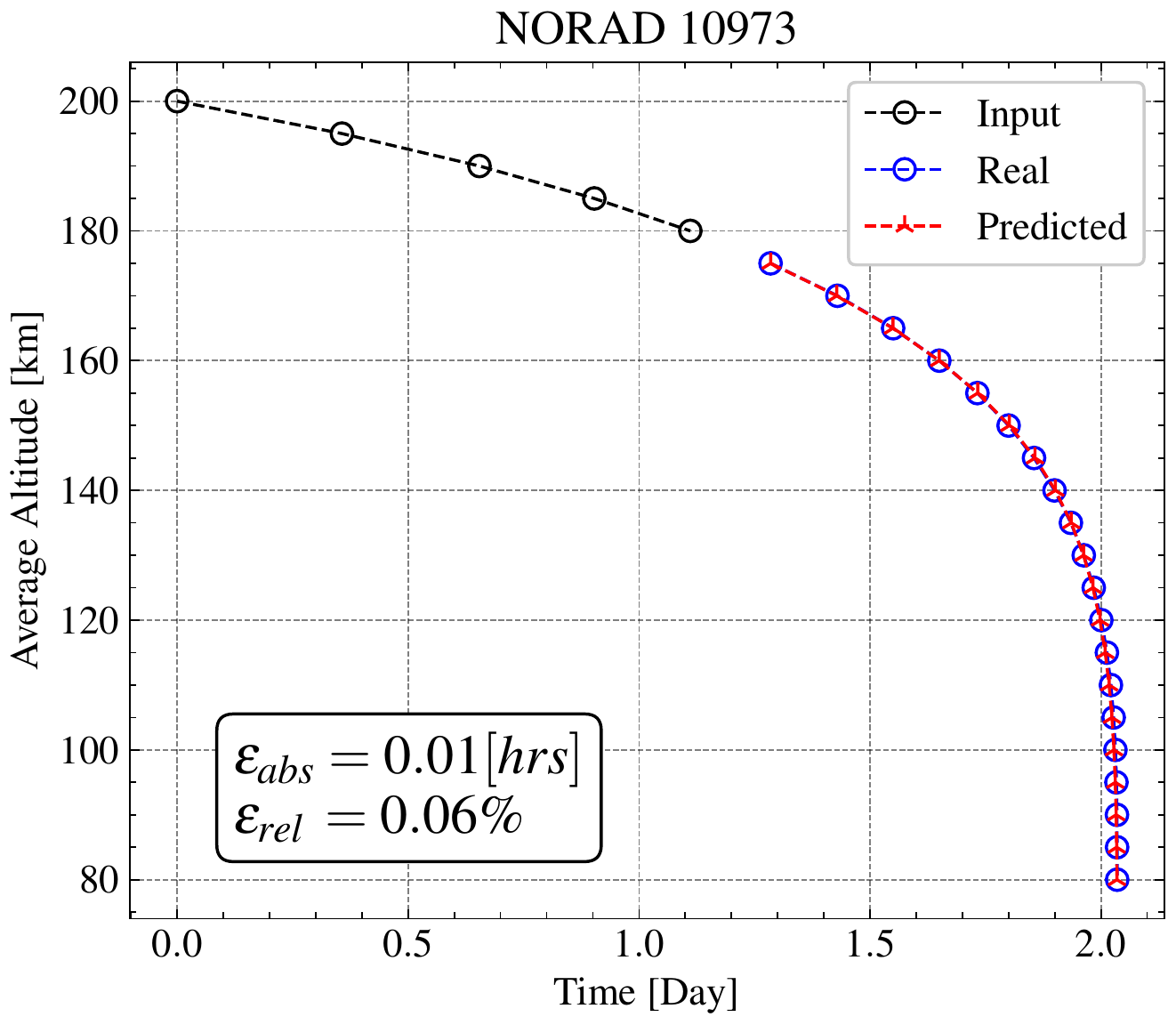}
		\caption{NORAD 10973}
	\end{subfigure}
	\hfill
	\begin{subfigure}{0.48\textwidth}
		\centering
		\label{fig:21701_A}
		\includegraphics[width=\textwidth]{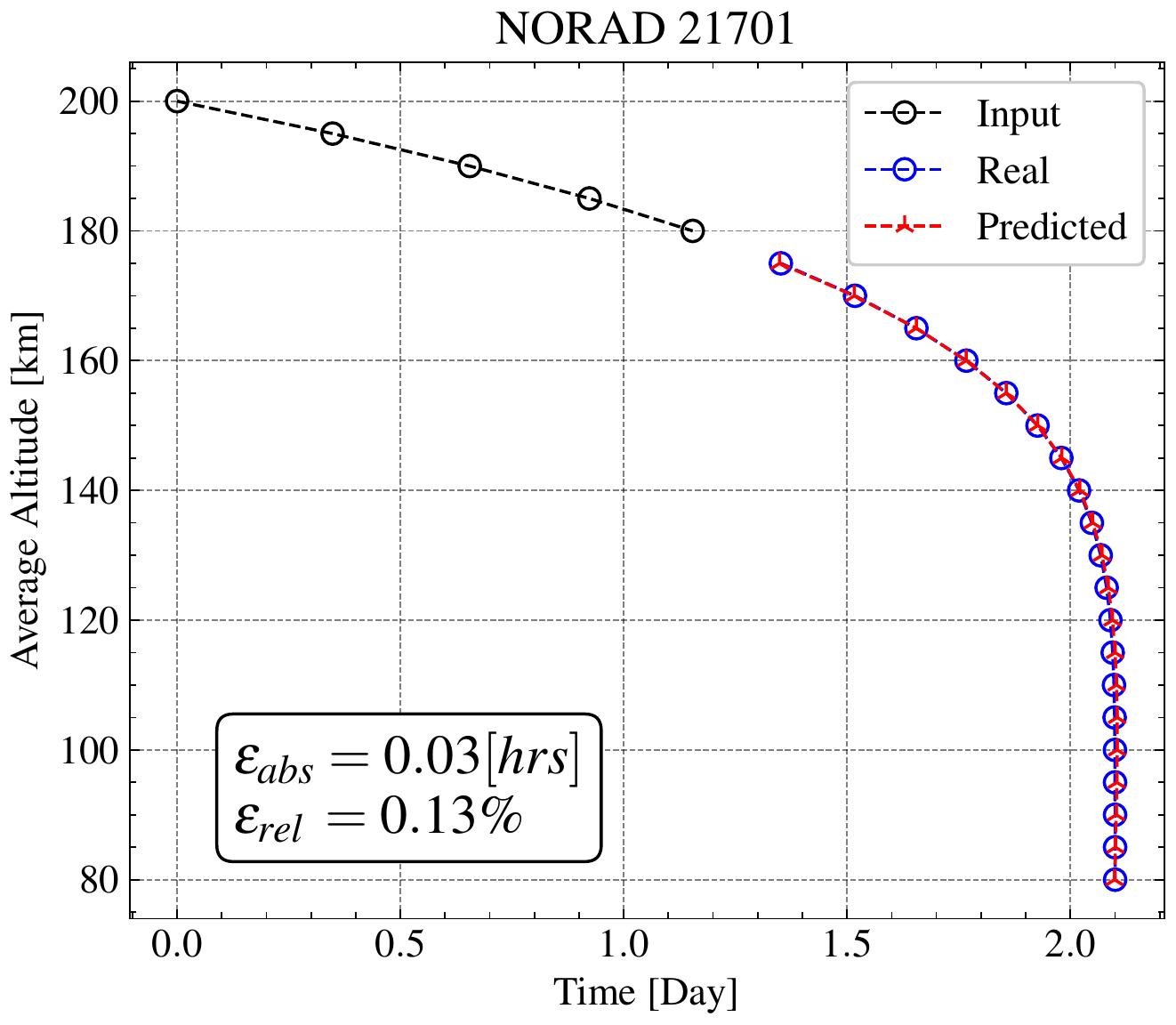}
		\caption{NORAD 21701}
	\end{subfigure}
	\\
	\vspace{0.25cm}
	\begin{subfigure}{0.48\textwidth}
		\centering
		\label{fig:31599_A}
		\includegraphics[width=\textwidth]{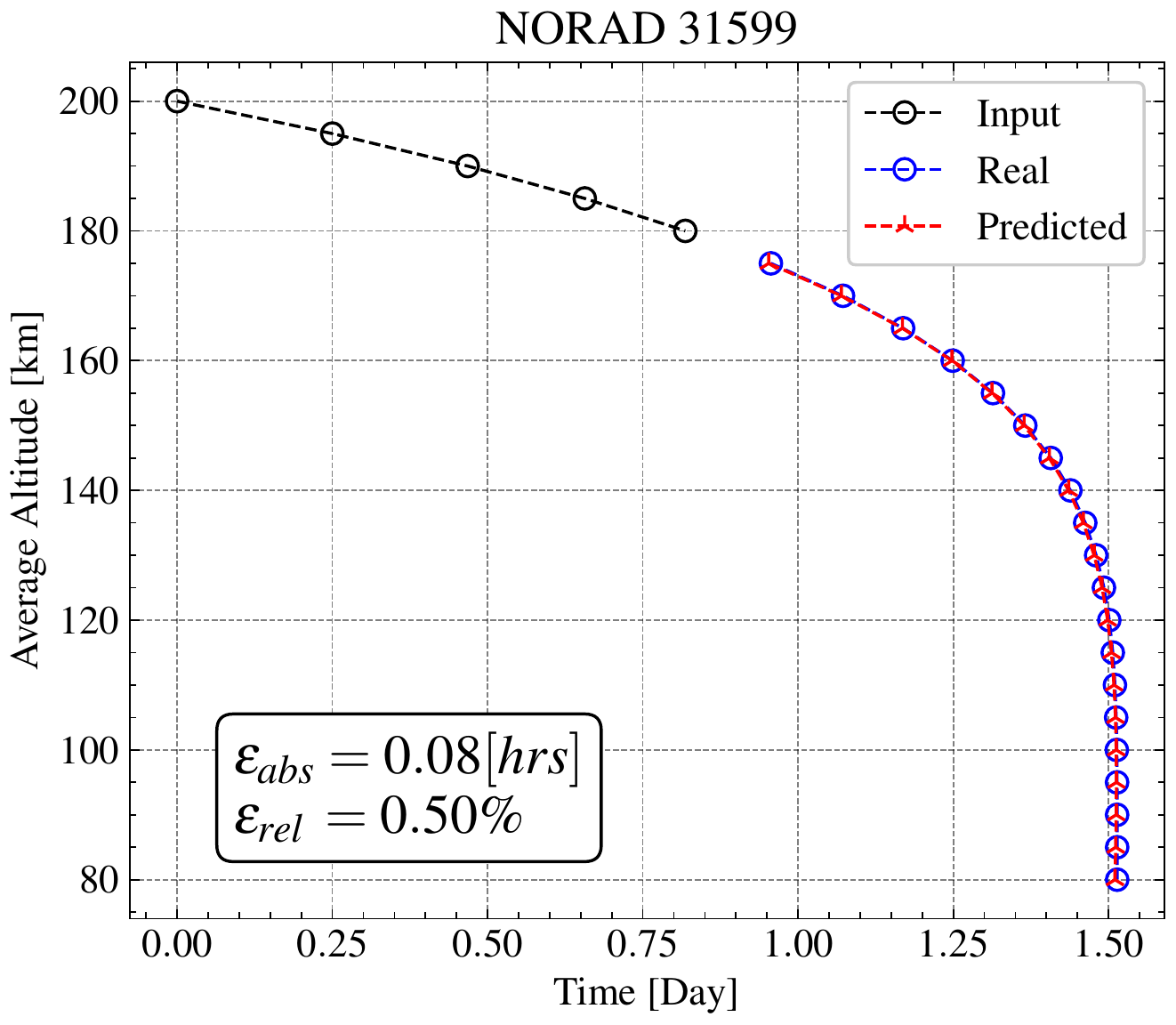}
		\caption{NORAD 31599}
	\end{subfigure}
	\hfill
	\begin{subfigure}{0.48\textwidth}
		\centering
		\label{fig:38086_A}
		\includegraphics[width=\textwidth]{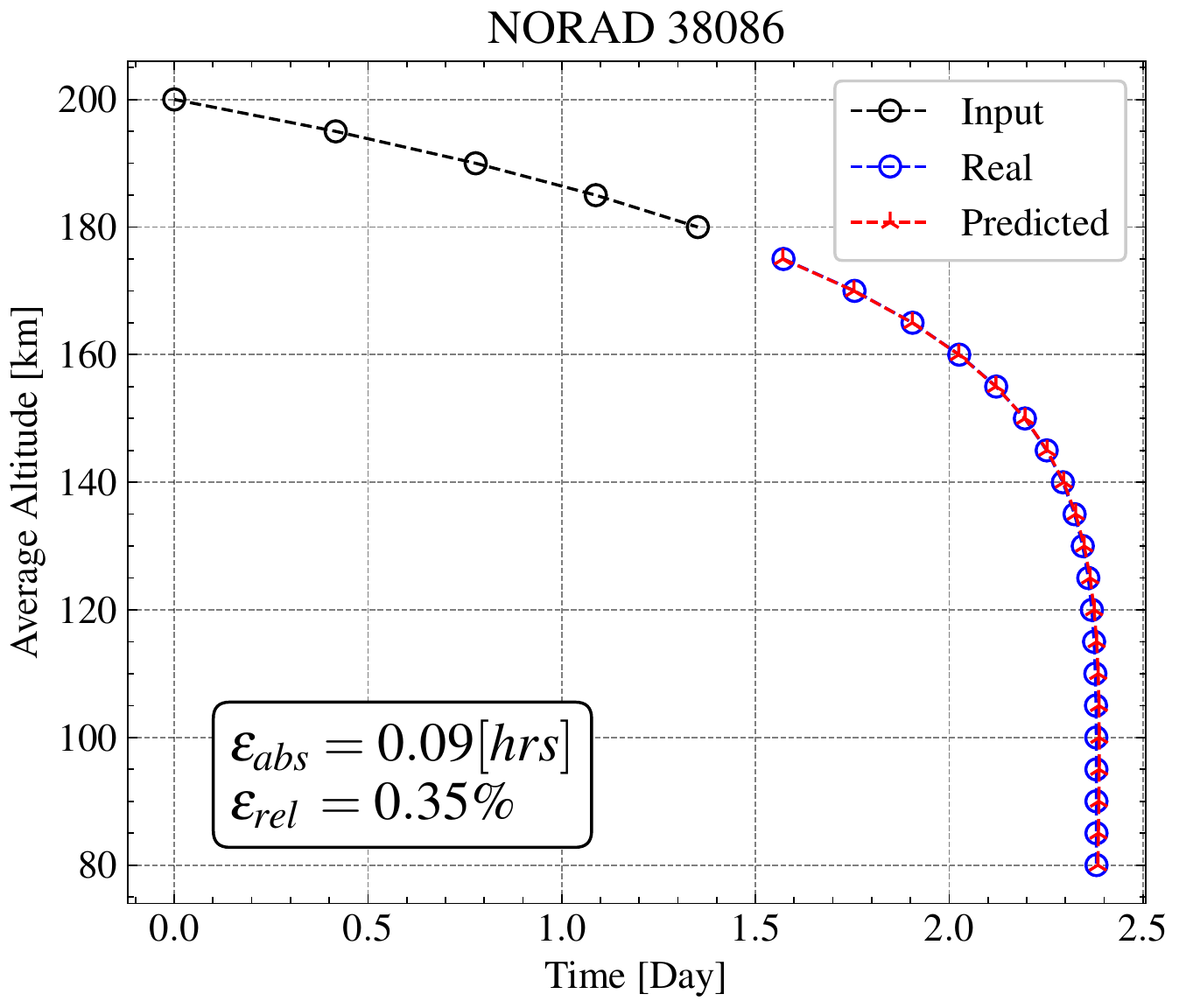}
		\caption{NORAD 38086}
	\end{subfigure}
	\\
	\vspace{0.25cm}
	\begin{subfigure}[b]{0.48\textwidth}
		\centering
	  \label{fig:39000_A}
	  \includegraphics[width=\textwidth]{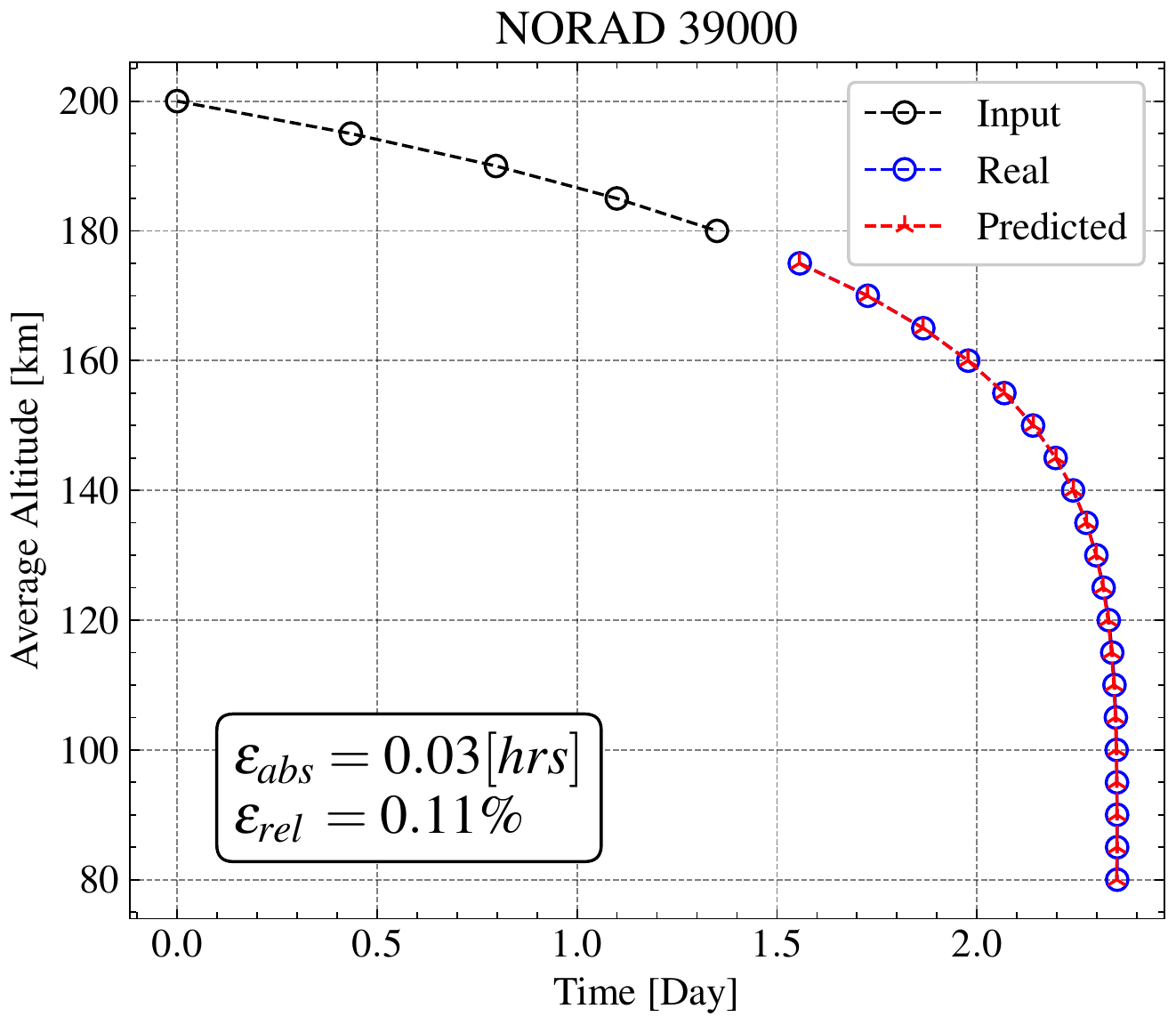}
	  \caption{NORAD 39000}
	\end{subfigure}
	\hfill
	\begin{subfigure}[b]{0.48\textwidth}
	  \centering
	  \label{fig:40138_A}
	  \includegraphics[width=\textwidth]{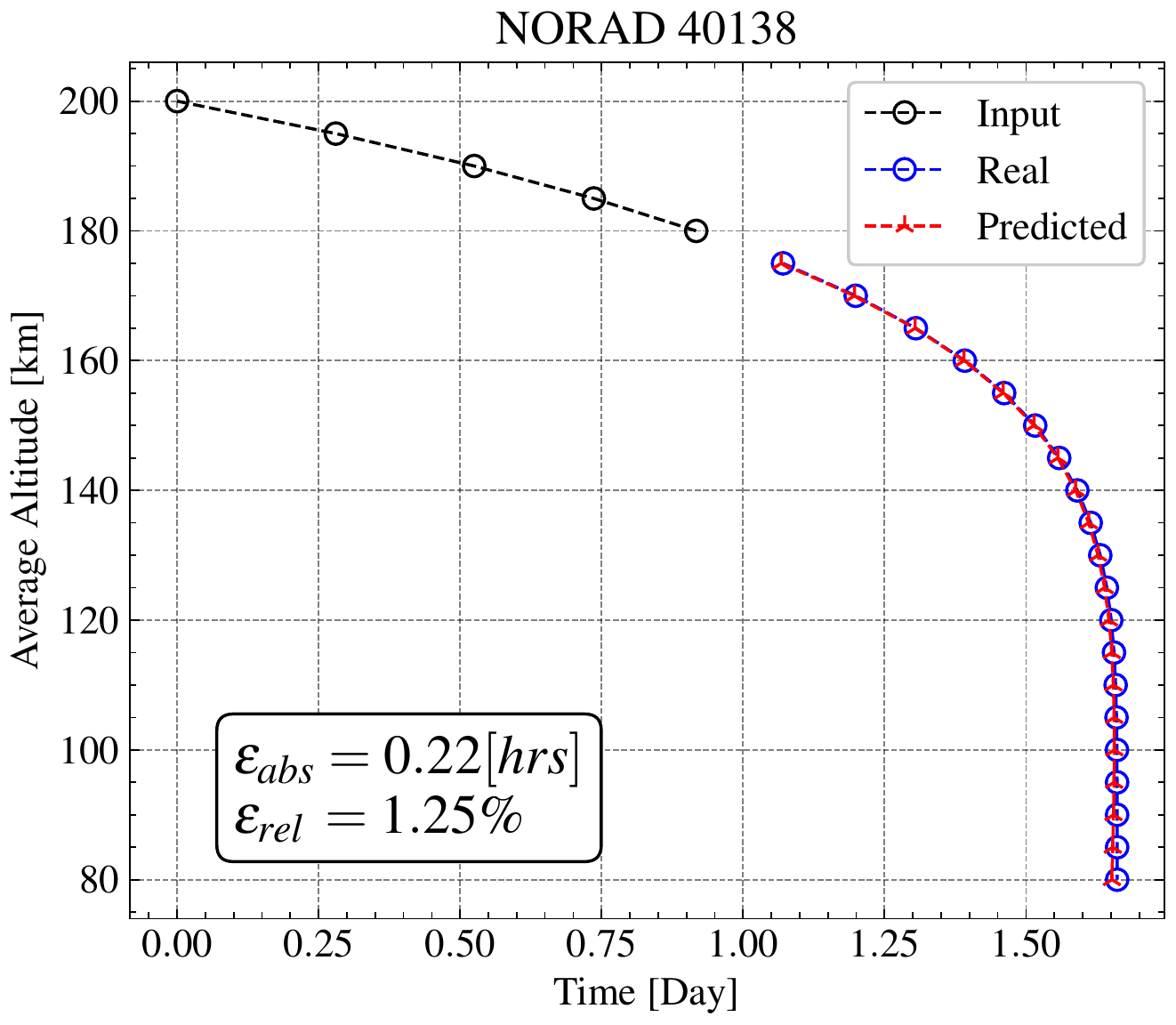}
	  \caption{NORAD 40138}
	\end{subfigure}
    \caption{Predictions of selected objects in category 1 from 180 km altitude. \label{fig:case_A_cat_1}}
\end{figure}

\cref{fig:case_A_cat_1} shows the trajectory prediction for selected objects of Category 1 in the test set. The black curve represents the input average altitude (the initial 5 elements of the input sequence, $T_x$), the red line represents the predicted sequence, and the blue trajectory defines the true output (as obtained from TLE data).

\begin{figure}[!htb]
    \centering
     \begin{subfigure}[b]{0.48\textwidth}
         \centering
         \includegraphics[width=\textwidth]{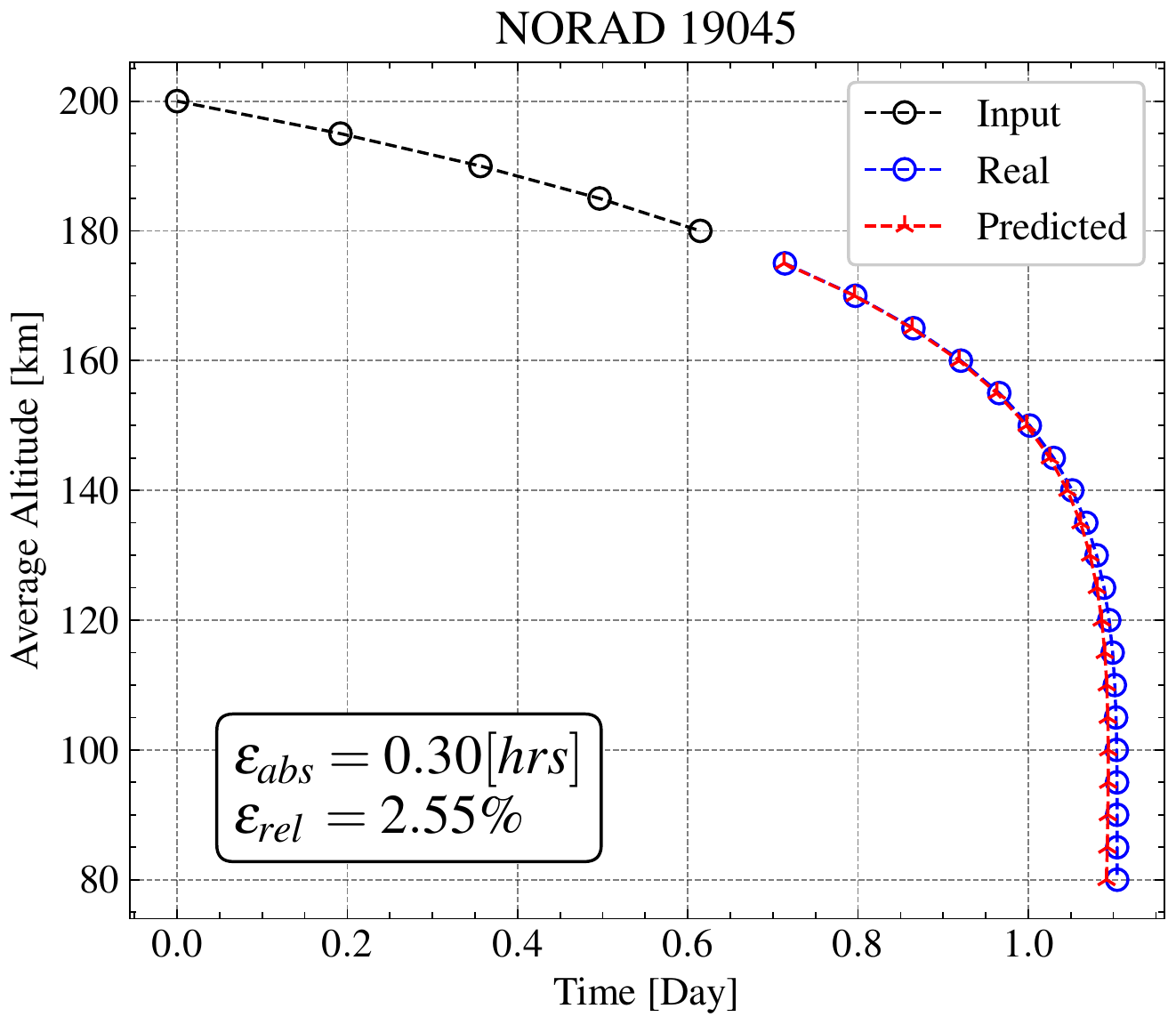}
         \caption{NORAD 19045.}
         \label{fig:19045_A}
     \end{subfigure}
     \hfill
     \begin{subfigure}[b]{0.48\textwidth}
         \centering
         \includegraphics[width=\textwidth]{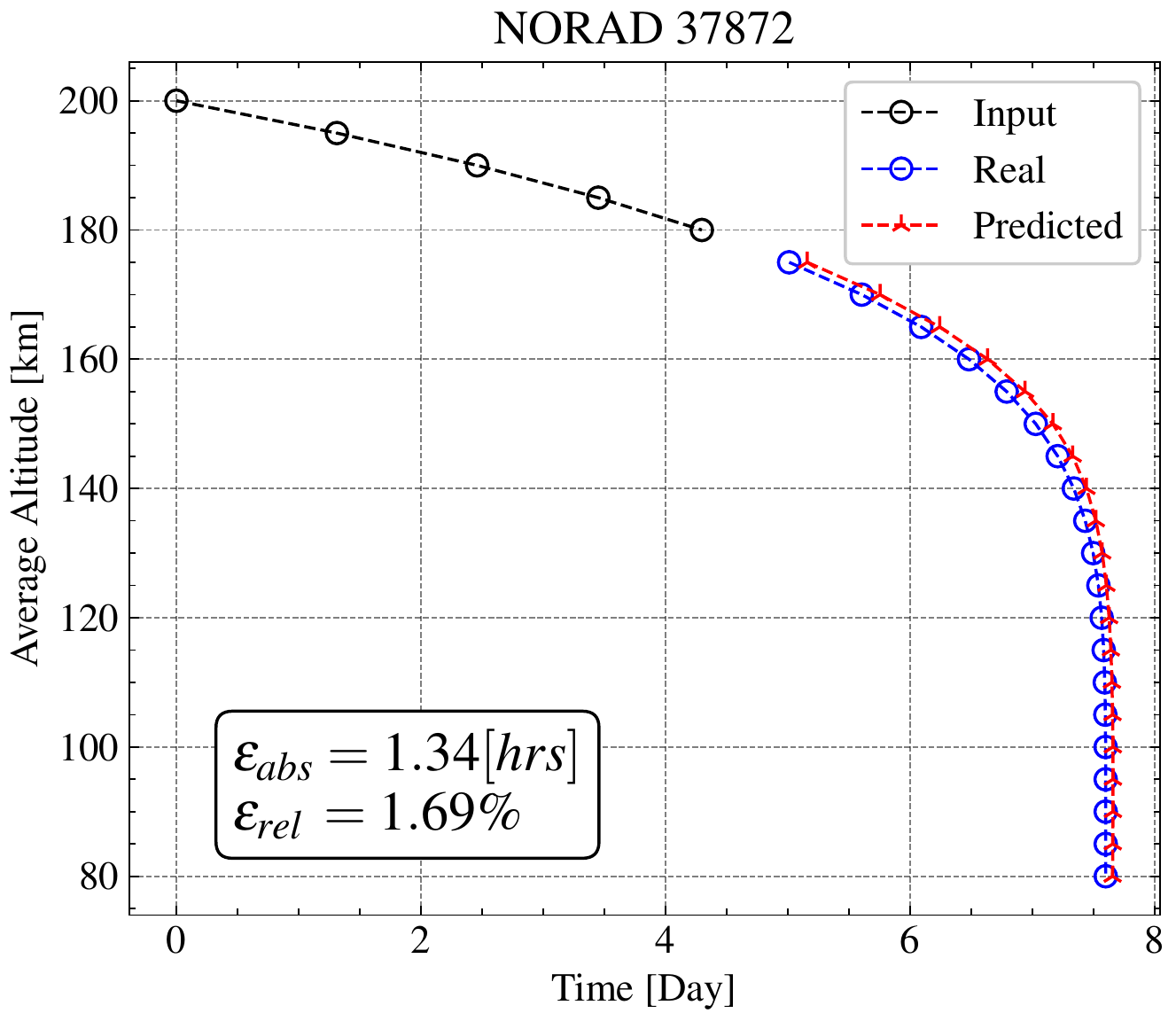}
         \caption{NORAD 37872.}
         \label{fig:37872_A}
     \end{subfigure}
    \caption{Predictions of 19045 and 37872 from 180 km altitude.}
    \label{fig:case_A_cat_2}
\end{figure}

\begin{figure}[!htb]
    \centering
     \begin{subfigure}[b]{0.48\textwidth}
         \centering
         \includegraphics[width=\textwidth]{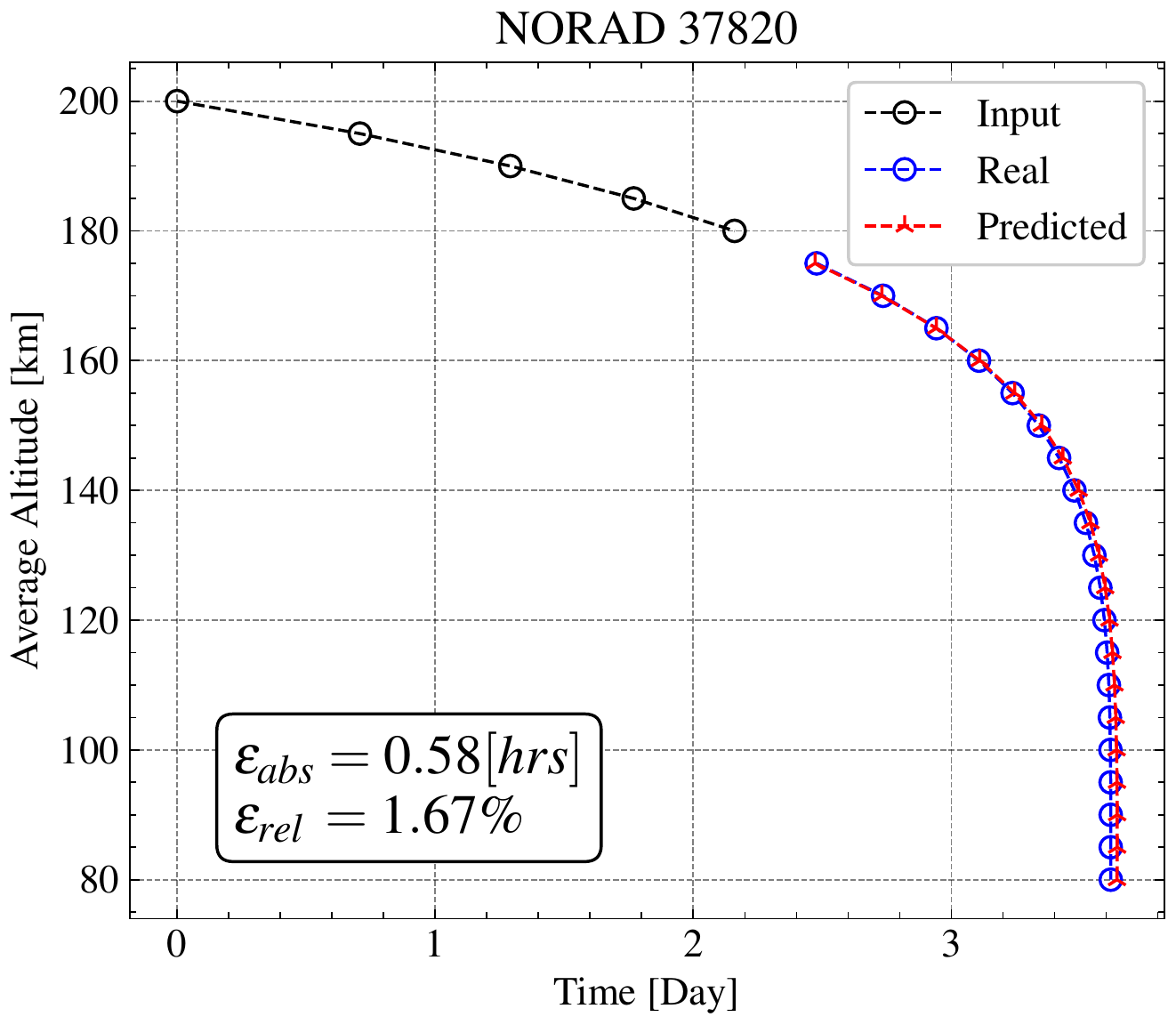}
         \caption{NORAD 37820.}
         \label{fig:37820_A}
     \end{subfigure}
     \hfill
     \begin{subfigure}[b]{0.48\textwidth}
         \centering
         \includegraphics[width=\textwidth]{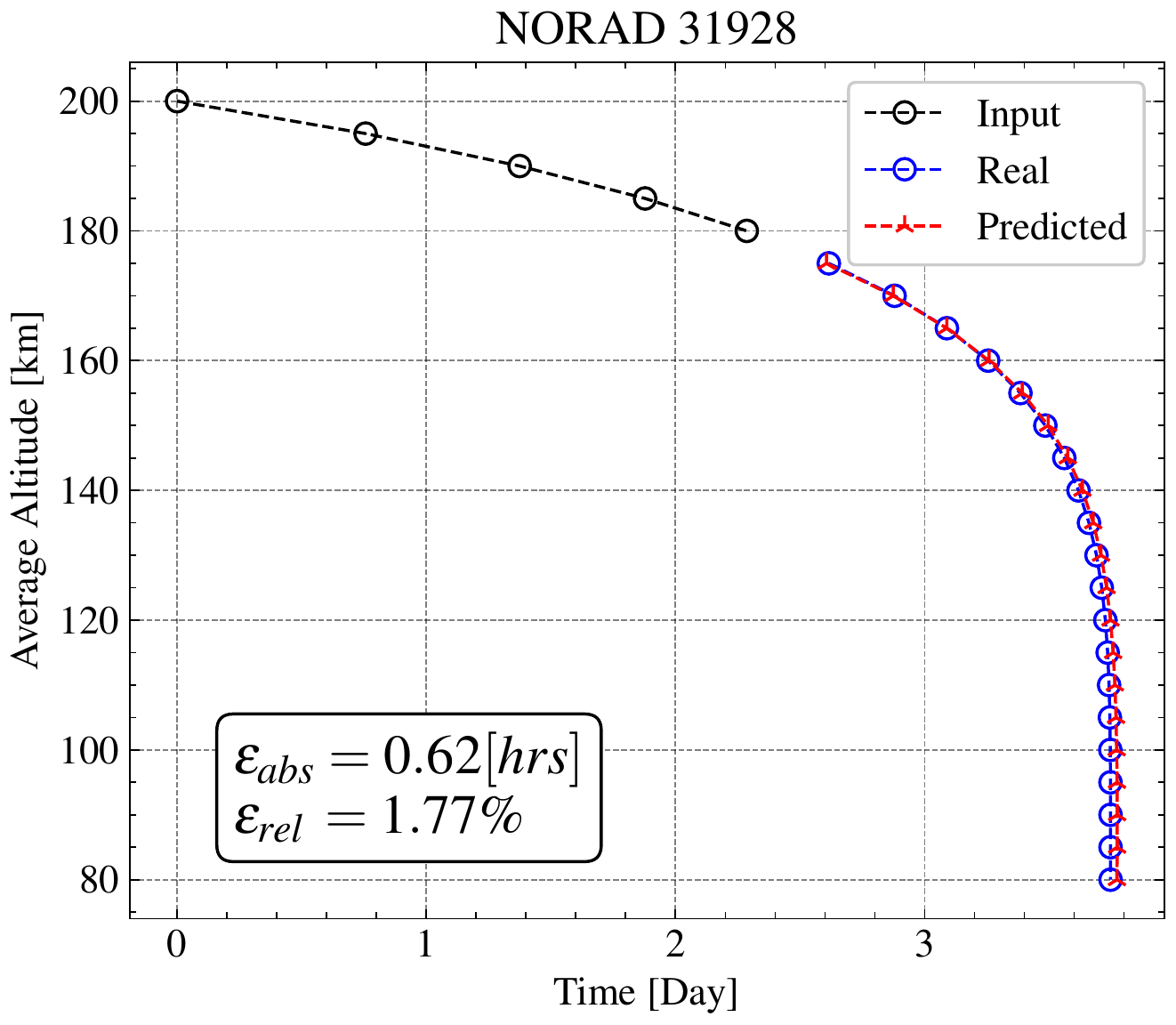}
         \caption{NORAD 31928.}
         \label{fig:31928_A}
     \end{subfigure}
    \caption{Predictions of 37820 and 31928 from 180 km altitude.}
    \label{fig:case_A_jung}
\end{figure}

In general, it can be seen that the model is capable of predicting the re-entry epoch with excellent results. The model can successfully estimate the output trajectory of each object because the outputs follow the true paths quite accurately. Indeed, as also shown in \cref{tab:results_cat1}, the highest MSE is associated with 38086; however, the relative errors on the re-entry epoch are small. Therefore, even if the MSE appears to be relatively high, the absolute and relative errors on the re-entry epoch are small. Moreover, the absolute errors are all lower than approximately 13 minutes, except for 40138, which is characterised by an error of 13.35 minutes. This can be explained by the higher $B^*$ distribution of 40138 with respect to the training set. Furthermore, considering objects 40138 and 10973, it appears that the model has correctly assimilated the knowledge related to the relation between the solar index and the drag-like coefficient.
\bigbreak
Concerning the objects with $B^*$ distributions different from the one of the training set (Category 2), the results are summarised in \cref{fig:case_A_cat_2} and \cref{fig:case_A_jung}.

\begin{figure}[!htb]
    \centering
    \includegraphics[width=0.7\columnwidth]{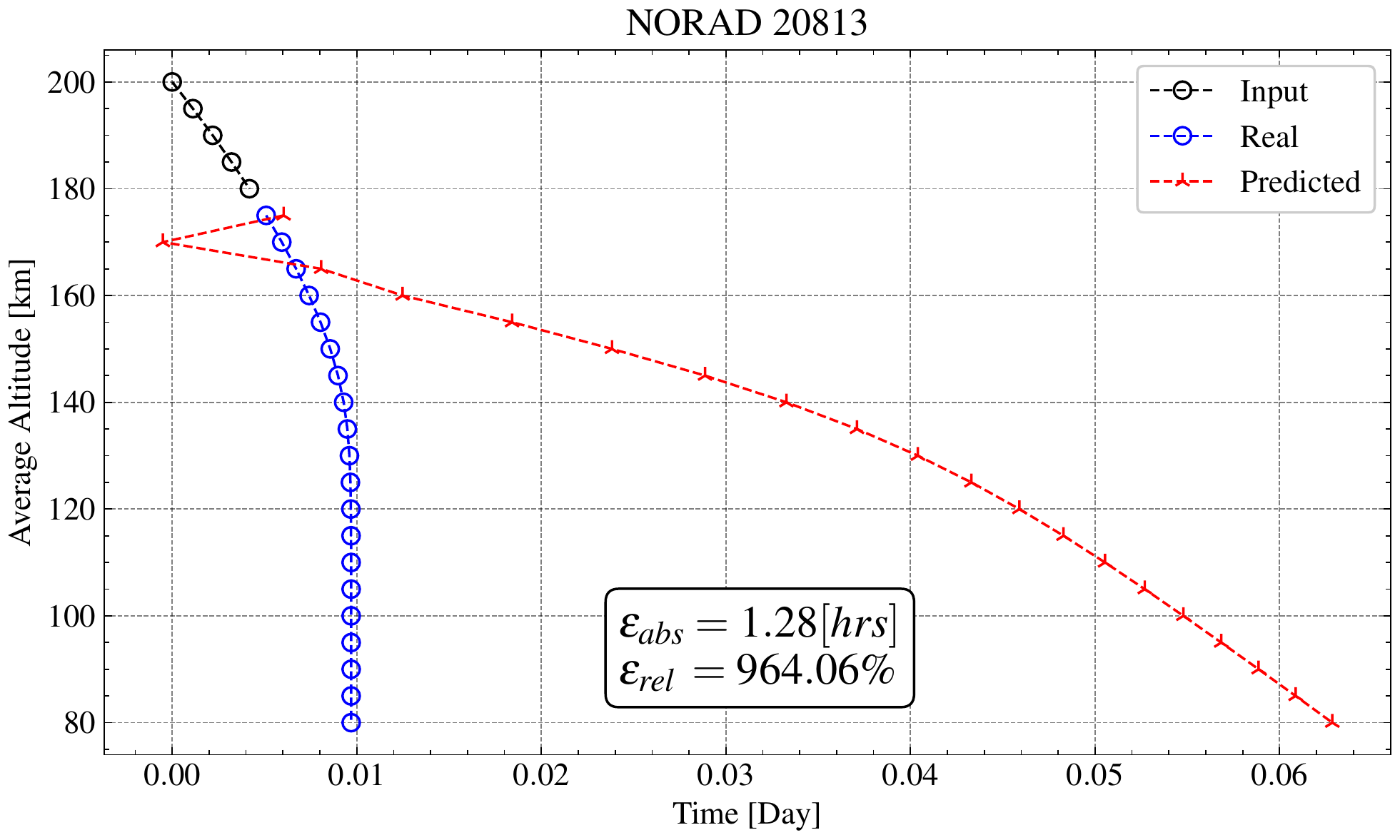}
    \caption{Prediction of NORAD 20813 from 180 km altitude.}
    \label{fig:20813_A}
\end{figure}

Generally, it can be observed that all errors are higher with respect to the previous case. This is related to the low number of objects characterised by a $B^*$ distribution similar to the training, which makes the model perform less accurately. However, in 4 out of 7 cases, the absolute errors are lower than approximately 40 minutes. Therefore, it can be concluded, in this case, that the model is capable of approximating the true output reasonably well. Nevertheless, object 20813 represents an exception in terms of errors, as can be seen in \cref{fig:20813_A}, where the output trajectory is completely distorted. 

By inspecting \cref{fig:boxplot_test_training}, it can be seen that the $B^*$ distribution of 20813 is characterised by a large range of values. However, its interquartile is lower than 19045,  and the model has proved itself to provide acceptable performances on the latter, with an output that accurately follows the true one. Therefore, it appears that the drag-like coefficient can not be directly linked to bad performance. Indeed, 20813 is characterised by high values of eccentricity, as highlighted in \cref{fig:box_plot_ecc}. This makes the nature of the re-entry different, and the model is not capable of acquiring this knowledge because it is not trained for it. Therefore, the prediction capabilities have to be related not only to the $B^*$ but also to the eccentricity.

\section{Conclusions and Discussion}
This work presents the development of a deep learning model that is based on a variant of the Seq2Seq architecture using curriculum learning for the re-entry prediction of uncontrolled objects in Low Earth Orbit based on TLE data. We therefore propose a paradigm shift from a physical to a data-driven approach. Based on the physical knowledge of the re-entry phenomenon, we introduced additional features alongside the average altitude profile, in an effort to better characterise and predict the re-entry epoch. Two additional key features have been identified: the drag-like coefficient, $B^*$, and the solar index, $\overline{F}_{10.7}$. The drag-like coefficient is of key importance as it relates the re-entry time to the interaction of the object with the atmosphere. In our work, we present a methodology to introduce such a parameter as a feature into a deep learning model, via a stepwise moving window interpolation scheme. The solar index is equivalently relevant in the re-entry prediction as it affects the density of the atmosphere and, with it, the re-entry time. Analysing this parameter, the need to introduce an additional feature has arisen, in order to relate the solar index with the physical properties of the object. Therefore, the area-to-mass ratio was also added as a feature. Studying the behaviour of the deep learning model with respect to these features, we identified possible ways to introduce them in the model.
The developed deep learning model has been tested on a set of objects studied during the IADC campaigns. It has been observed, in accordance with Jung et al. \cite{jung2021recurrent}, that the $B^*$ distribution has significant importance on the performances of the model. However, the poor performance related to object 20813 (\cref{fig:20813_A}) suggests that also eccentricity has a relevant impact. Hence, future works may include the processing of highly eccentricity orbits. In addition, it has been found that a decrease in the selected loss function (MSE) did not necessarily correspond to higher accuracy of the re-entry prediction. Consequently, future studies could also investigate the effect of different loss functions on the output of the deep learning model. Overall, the analyses showed promising results for the re-entry prediction using machine learning; however, as can be expected, the prediction capabilities of the model strongly depend on the quality and quantity of the training data. It is therefore crucial, in order to expand the capabilities of this data-driven approach, to expand the size of the training data and, potentially, improve the quality of such data by considering additional sources other than TLEs.

\vspace{6pt} 



\authorcontributions{Conceptualization, Mirko Trisolini, Francesco Salmaso, Camilla Colombo; methodology, Francesco Salmaso, Mirko Trisolini; software, Francesco Salmaso; validation, Francesco Salmaso; formal analysis, Francesco Salmaso; investigation, Francesco Salmaso; writing---original draft preparation, Mirko Trisolini, Francesco Salmaso; writing---review and editing, Mirko Trisolini, Camilla Colombo; visualization, Francesco Salmaso; supervision, Mirko Trisolini, Camilla Colombo; funding acquisition, Camilla Colombo. All authors have read and agreed to the published version of the manuscript.}

\funding{The research presented in this thesis received funding from the European Research Council (ERC) under the European Union’s Horizon 2020 research and innovation programme as part of the COMPASS project (Grant agreement No 679086).}



\conflictsofinterest{The authors declare no conflict of interest.} 

\abbreviations{Abbreviations}{
The following abbreviations are used in this manuscript:\\

\noindent 
\begin{tabular}{@{}ll}
TLE & Two Line Element \\
TIP & Tracking and Impact Prediction \\
PCA & Principal Component Analysis \\
Seq2Seq & Sequence to Sequence \\
TPE & Tree-structured Parzen Estimator \\
ASHA & Asynchronous Successive Halving Algorithm \\
ANN & Artificial Neural Network \\
RNN & Recurrent Neural Network \\
GRU & Gated Recurrent Unit \\
MSE & Mean Squared Error \\
OMM & Orbital Mean-Elements Message \\
\end{tabular}
}

\appendixtitles{no} 
\appendixstart
\appendix
\section[\appendixname~\thesection]{}

\label{ch:appendix_a}
\subsection{NORAD IDs of training set}
\label{sssec:norad_val}
\noindent 22277, 33594, 38223, 43538, 49130, 38976, 43033,
37727, 12389, 28883, 45942, 45729, 33440, 40459,
28824, 45601, 39573, 41813, 25431, 29504, 25577,
26481, 12072, 40714, 40700, 33448, 41640, 43703,
40120, 43586, 1377, 37346, 28701, 44500, 36604,
39262, 28642, 38349, 28842, 33387, 40085, 28827,
39561, 24968, 28400, 49045, 28822, 34670, 24840,
38048, 25288, 33592, 48853, 34906, 28486, 28424,
29480, 41478, 39259, 39683, 25274, 27451, 39147,
28641, 25291, 3835, 46930, 39507, 27375, 24837,
31602, 10861, 46390, 25285, 28655, 40949, 40452,
41313, 38075, 8063, 41628, 28504, 12071, 43705,
43475, 25173, 22781, 24872, 40589, 42718, 39145,
40897, 43007, 29058, 33507, 40096, 28896, 40728,
44210, 28625, 32267, 37674, 37397, 39033, 37874,
39566, 28777, 25064, 11745, 34872, 40957, 43243,
25342, 44228, 32007, 39409, 32377, 42701, 44070,
36835, 41865, 46397, 36096, 27372, 39178, 26897,
42734, 29247, 39116, 33456, 43166, 32261, 40745,
25647, 27474, 23769, 40421, 39560, 40430, 40740,
25275, 35642, 41821, 27450, 31596, 26874, 49065,
44551, 41819, 44385, 27374, 48804, 32485, 40723,
40453, 25290, 41100, 32269, 39193, 24945, 37883,
37821, 43532, 29053, 40314, 39580, 41559, 28933,
40668, 40945, 32270, 46614, 43494, 27642, 28823,
28362, 41480, 43757, 14207, 29157, 39571, 33054,
41776, 28425, 40363, 28878, 37197, 29109, 41909,
28519, 38848, 39503, 32757, 37172, 12054, 34603,
21694, 42723, 43863, 33435, 45112, 25468, 39527,
40127, 45596, 28099, 11474, 29667, 40702, 38462,
10582, 38074, 11849, 14694, 27392, 43691, 44372,
36130, 21931, 38249, 25276, 39623, 25040, 25106,
32477, 30778, 29080, 42757, 41486, 39519, 32262,
37860, 27841, 43090, 31394, 28867, 44844, 43239,
32002, 39126, 44707, 35694, 41596, 37680, 41769,
25171, 23020, 39514, 29229, 29263, 42731, 38862,
32751, 29253, 47619, 44438, 32059, 25287, 39393,
20299, 27700, 43212, 35818, 39031, 38037, 35867,
29403, 41574, 28363, 33457, 44206, 24794, 37255,
44834, 41777, 40727, 27550, 28907, 37217, 35947,
37942, 29653, 43667, 39569, 39513, 25777, 7338,
39062, 35011, 36362, 41125, 37634, 42703, 43244,
29394, 40898, 40126, 43637, 44826, 39220, 42972,
25263, 41027, 39171, 24839, 45604, 36506, 27706,
24905, 41487, 37758, 43027, 46669, 28403, 25530,
41572, 44799, 29386, 39180, 39149, 41597, 47255,
28130, 29246, 39386, 37878, 47348, 42730, 41387,
41178, 35941, 38672, 24966\\

\subsection{NORAD IDs of validation Set}
\noindent 41868, 45177, 29488, 39682, 60,
32386, 38336, 33332, 38550,
24792, 48160, 36522, 32284,
39649, 35949, 43064, 41636,
28191, 40247, 6212, 25346,
37362, 11671, 28643, 44317,
42938, 24904, 43129, 11332,
28445, 27475, 41107, 40393,
44505, 40313, 36512, 48866,
36087, 37873, 37729, 42716,
39363, 48870, 44111, 40359,
13402, 39187, 33273, 47782,
26405, 37856, 28506, 41455,
39408, 39493, 41475, 39137,
24869, 34840, 42733, 41629,
48275, 23757, 46463, 40361,
40619, 28997, 42702, 42800,
28942, 44075, 36749, 32713,
33449, 38076, 37876, 41449,
28774, 31798, 41483, 45938,
39532, 37383, 40886
 
\subsection{NORAD IDs of test set}
\noindent 20813, 19045, 11601, 10973, 40138, 38086, 23853, 21701, 37872, 39000, 20638, 31928, 37820, 26873, 31599

\subsubsection{Test set subdivision in Category 1 and Category 2}
\label{subsubsec:test_obj_categories}

\cref{tab:division_test_bstar} shows the NORAD IDs of the test set objects subdivided into Category 1 and Category 2. Category 1 contains objects with a median $B^{*}$ within the interquartile distance of the $B^{*}$ distribution of the training set, while objects in Category 2 have a median $B^{*}$ outside this interval. 

\begin{table}[!htb]
\centering
\begin{tabular}{c c c c c c c c c c}
\hline 
\hline
\textbf{Category}       &  \multicolumn{9}{c}{\textbf{NORAD IDs}}\\\\
\hline
\textbf{1}    &  10973 & 11601 & 20638 & 21701 & 26873 & 31599 & 38086 & 39000 & 40138 \\
\hline
\textbf{2}    &  19045 & 20813 & 23853 & 31928 & 37820 & 37872 &       &       &\\
\hline
\hline
\end{tabular}
\caption{Subdivision of the bodies in the test set according to the $B^*$ distributions.}
\label{tab:division_test_bstar}
\end{table}

\begin{adjustwidth}{-\extralength}{0cm}

\reftitle{References}


\bibliography{bibliography}

\begin{thebibliography}{999}

\bibitem[{ESA Space Debris Office}(2021)]{esa_report}
{ESA Space Debris Office}.
\newblock {ESA}’S {A}nnual {S}pace {E}nvironent {R}eport,  2021.

\bibitem[Alby \em{et~al.}(2004)Alby, Alwes, Anselmo, Baccini, Bonnal, Crowther,
  Flury, Jehn, Klinkrad, Portelli, and Tremayne-Smith]{ALBY20041260}
Alby, F.; Alwes, D.; Anselmo, L.; Baccini, H.; Bonnal, C.; Crowther, R.; Flury,
  W.; Jehn, R.; Klinkrad, H.; Portelli, C.;  et~al.
\newblock The {E}uropean {S}pace {D}ebris {S}afety and {M}itigation {S}tandard.
\newblock {\em Advances in Space Research} {\bf 2004}, {\em 34},~1260--1263.
\newblock {\url{https://doi.org/10.1016/j.asr.2003.08.043}}.

\bibitem[{NASA}(2019)]{nasa_debris}
{NASA}.
\newblock {Process for Limiting Orbital Debris, NASA-STD-8719.14B},  2019.

\bibitem[spa()]{space_track}
Space-{T}rack.
\newblock www.space-track.org.

\bibitem[Pardini and Anselmo(2013)]{results_uncontrolled}
Pardini, C.; Anselmo, L.
\newblock Re-entry predictions for uncontrolled satellites: results and
  challenges.
\newblock In Proceedings of the 6th IAASS Conference "Safety is Not an Option";
  ,  2013.

\bibitem[Pardini and Anselmo(2003)]{pardini2003performance}
Pardini, C.; Anselmo, L.
\newblock Performance evaluation of atmospheric density models for satellite
  reentry predictions with high solar activity levels.
\newblock {\em Transactions of The Japan Society for Aeronautical and Space
  Sciences} {\bf 2003}, {\em 46},~42--46.
\newblock {\url{https://doi.org/10.2322/tjsass.46.42}}.

\bibitem[Pardini and Anselmo(2018)]{PARDINI201846}
Pardini, C.; Anselmo, L.
\newblock Assessing the risk and the uncertainty affecting the uncontrolled
  re-entry of manmade space objects.
\newblock {\em Journal of Space Safety Engineering} {\bf 2018}, {\em
  5},~46--62.
\newblock {\url{https://doi.org/10.1016/j.jsse.2018.01.003}}.

\bibitem[Braun \em{et~al.}(2012)Braun, Flegel, Gelhaus, Kebschull, Moeckel,
  Wiedemann, Sánchez-Ortiz, Krag, and Vörsmann]{solarmodel}
Braun, V.; Flegel, S.; Gelhaus, J.; Kebschull, C.; Moeckel, M.; Wiedemann, C.;
  Sánchez-Ortiz, N.; Krag, H.; Vörsmann, P.
\newblock Impact of Solar Flux Modeling on Satellite Lifetime Predictions.
\newblock In Proceedings of the 63rd International Astronautical Congress; ,
  2012.

\bibitem[Vallado and Finkleman(2014)]{VALLADO2014141}
Vallado, D.A.; Finkleman, D.
\newblock A critical assessment of satellite drag and atmospheric density
  modeling.
\newblock {\em Acta Astronautica} {\bf 2014}, {\em 95},~141--165.
\newblock {\url{https://doi.org/10.1016/j.actaastro.2013.10.005}}.

\bibitem[Vallado(2001)]{vallado2001fundamentals}
Vallado, D.
\newblock {\em Fundamentals of {A}strodynamics and {A}pplications}, second ed.;
  Springer Dordrecht,  2001.

\bibitem[Anselmo and Pardini(2005)]{anselmo2005computational}
Anselmo, L.; Pardini, C.
\newblock Computational methods for reentry trajectories and risk assessment.
\newblock {\em Advances in Space Research} {\bf 2005}, {\em 35},~1343--1352.
\newblock {\url{https://doi.org/10.1016/j.asr.2005.04.089}}.

\bibitem[Frey \em{et~al.}(2019)Frey, Colombo, and Lemmens]{FREY20191}
Frey, S.; Colombo, C.; Lemmens, S.
\newblock Extension of the King-Hele orbit contraction method for accurate,
  semi-analytical propagation of non-circular orbits.
\newblock {\em Advances in Space Research} {\bf 2019}, {\em 64},~1--17.
\newblock {\url{https://doi.org/https://doi.org/10.1016/j.asr.2019.03.016}}.

\bibitem[Jung \em{et~al.}(2021)Jung, Seong, Jung, and Bang]{jung2021recurrent}
Jung, O.; Seong, J.; Jung, Y.; Bang, H.
\newblock Recurrent neural network model to predict re-entry trajectories of
  uncontrolled space objects.
\newblock {\em Advances in Space Research} {\bf 2021}, {\em 68},~2515--2529.
\newblock {\url{https://doi.org/10.1016/j.asr.2021.04.041}}.

\bibitem[Lidtke \em{et~al.}(2019)Lidtke, Gondelach, and
  Armellin]{LIDTKE20191289}
Lidtke, A.A.; Gondelach, D.J.; Armellin, R.
\newblock Optimising filtering of two-line element sets to increase re-entry
  prediction accuracy for GTO objects.
\newblock {\em Advances in Space Research} {\bf 2019}, {\em 63},~1289--1317.
\newblock {\url{https://doi.org/10.1016/j.asr.2018.10.018}}.

\bibitem[Flohrer \em{et~al.}(2008)Flohrer, Krag, and
  Klinkrad]{flohrer2008assessment}
Flohrer, T.; Krag, H.; Klinkrad, H.
\newblock Assessment and categorization of TLE orbit errors for the US SSN
  catalogue.
\newblock In Proceedings of the Advanced Maui Optical and Space Surveillance
  Technologies (AMOS) Conference,  2008.

\bibitem[Levit and Marshall(2011)]{levit2011improved}
Levit, C.; Marshall, W.
\newblock Improved orbit predictions using two-line elements.
\newblock {\em Advances in Space Research} {\bf 2011}, {\em 47},~1107--1115.
\newblock {\url{https://doi.org/10.1016/j.asr.2010.10.017}}.

\bibitem[Aida and Kirschner(2013)]{aida2013accuracy}
Aida, S.; Kirschner, M.
\newblock Accuracy assessment of SGP4 orbit information conversion into
  osculating elements.
\newblock In Proceedings of the 6th European Conference on Space Debris; ,
  2013.

\bibitem[Raschka(2019)]{raschka2015python}
Raschka, S.
\newblock {\em Python machine learning}, third ed.; Packt Publishing Ltd,
  2019.

\bibitem[Sutskever \em{et~al.}(2014)Sutskever, Vinyals, and
  Le]{sutskever2014seq2seq}
Sutskever, I.; Vinyals, O.; Le, Q.V.
\newblock Sequence to Sequence Learning with Neural Networks.
\newblock In Proceedings of the 27th International Conference on Neural
  Information Processing Systems - Volume 2; MIT Press: Cambridge, MA, USA,
  2014; p. 3104–3112.

\bibitem[Zhang \em{et~al.}(2021)Zhang, Lipton, Li, and Smola]{zhang2021dive}
Zhang, A.; Lipton, Z.C.; Li, M.; Smola, A.J.
\newblock Dive into Deep Learning,  2021.
\newblock arXiv:2106.11342.

\bibitem[Goodfellow \em{et~al.}(2016)Goodfellow, Bengio, and
  Courville]{goodfellow2016deep}
Goodfellow, I.; Bengio, Y.; Courville, A.
\newblock {\em Deep learning}; MIT press,  2016.

\bibitem[Elman(1990)]{elman1990finding}
Elman, J.L.
\newblock Finding structure in time.
\newblock {\em Cognitive science} {\bf 1990}, {\em 14},~179--211.
\newblock {\url{https://doi.org/10.1207/s15516709cog1402_1}}.

\bibitem[Cho \em{et~al.}(2014)Cho, van Merri{\"e}nboer, Gulcehre, Bahdanau,
  Bougares, Schwenk, and Bengio]{cho-etal-2014-learning}
Cho, K.; van Merri{\"e}nboer, B.; Gulcehre, C.; Bahdanau, D.; Bougares, F.;
  Schwenk, H.; Bengio, Y.
\newblock Learning Phrase Representations using {RNN} Encoder{--}Decoder for
  Statistical Machine Translation.
\newblock In Proceedings of the 2014 Conference on Empirical Methods in Natural
  Language Processing ({EMNLP}); Association for Computational Linguistics:
  Doha, Qatar,  2014; pp. 1724--1734.
\newblock {\url{https://doi.org/10.3115/v1/D14-1179}}.

\bibitem[Bengio \em{et~al.}(2015)Bengio, Vinyals, Jaitly, and
  Shazeer]{bengio2015scheduled}
Bengio, S.; Vinyals, O.; Jaitly, N.; Shazeer, N.
\newblock Scheduled sampling for sequence prediction with recurrent neural
  networks.
\newblock {\em Advances in neural information processing systems} {\bf 2015},
  {\em 28}.

\bibitem[Chung \em{et~al.}(2014)Chung, Gulcehre, Cho, and
  Bengio]{chung2014empirical}
Chung, J.; Gulcehre, C.; Cho, K.; Bengio, Y.
\newblock Empirical evaluation of gated recurrent neural networks on sequence
  modeling,  2014.
\newblock arXiv:1412.3555.

\bibitem[Abadi \em{et~al.}(2015)Abadi, Agarwal, Barham, Brevdo, Chen, Citro,
  Corrado, Davis, Dean, Devin, Ghemawat, Goodfellow, Harp, Irving, Isard, Jia,
  Jozefowicz, Kaiser, Kudlur, Levenberg, Man\'{e}, Monga, Moore, Murray, Olah,
  Schuster, Shlens, Steiner, Sutskever, Talwar, Tucker, Vanhoucke, Vasudevan,
  Vi\'{e}gas, Vinyals, Warden, Wattenberg, Wicke, Yu, and
  Zheng]{tensorflow2015-whitepaper}
Abadi, M.; Agarwal, A.; Barham, P.; Brevdo, E.; Chen, Z.; Citro, C.; Corrado,
  G.S.; Davis, A.; Dean, J.; Devin, M.;  et~al.
\newblock {TensorFlow}: Large-Scale Machine Learning on Heterogeneous Systems,
  2015.
\newblock Software available from tensorflow.org.

\bibitem[Chollet(2021)]{chollet2021deep}
Chollet, F.
\newblock {\em Deep learning with Python}; Simon and Schuster,  2021.

\bibitem[Dong and Liu(2018)]{dong2018feature}
Dong, G.; Liu, H.
\newblock {\em Feature engineering for machine learning and data analytics};
  CRC Press,  2018.

\bibitem[Salmaso(2022)]{Salmaso2022thesis}
Salmaso, F.
\newblock Machine learning model for uncontrolled re-entry predictions of space
  objects and feature engineering.
\newblock PhD thesis, Politecnico di Milano,  2022.

\bibitem[Virtanen \em{et~al.}(2020)Virtanen, Gommers, Oliphant, Haberland,
  Reddy, Cournapeau, Burovski, Peterson, Weckesser, Bright, {van der Walt},
  Brett, Wilson, Millman, Mayorov, Nelson, Jones, Kern, Larson, Carey, Polat,
  Feng, Moore, {VanderPlas}, Laxalde, Perktold, Cimrman, Henriksen, Quintero,
  Harris, Archibald, Ribeiro, Pedregosa, {van Mulbregt}, and {SciPy 1.0
  Contributors}]{2020SciPy-NMeth}
Virtanen, P.; Gommers, R.; Oliphant, T.E.; Haberland, M.; Reddy, T.;
  Cournapeau, D.; Burovski, E.; Peterson, P.; Weckesser, W.; Bright, J.;
  et~al.
\newblock {{SciPy} 1.0: Fundamental Algorithms for Scientific Computing in
  Python}.
\newblock {\em Nature Methods} {\bf 2020}, {\em 17},~261--272.
\newblock {\url{https://doi.org/10.1038/s41592-019-0686-2}}.

\bibitem[Curtis(2014)]{curtis2013orbital}
Curtis, H.D.
\newblock {\em Orbital Mechanics for Engineering Students}, third ed.;
  Butterworth-Heinemann,  2014.

\bibitem[Bergstra \em{et~al.}(2013)Bergstra, Yamins, and
  Cox]{pmlr-v28-bergstra13}
Bergstra, J.; Yamins, D.; Cox, D.
\newblock Making a Science of Model Search: Hyperparameter Optimization in
  Hundreds of Dimensions for Vision Architectures.
\newblock In Proceedings of the Proceedings of the 30th International
  Conference on Machine Learning; ,  2013; Vol.~28, pp. 115--123.

\bibitem[Li \em{et~al.}(2020)Li, Jamieson, Rostamizadeh, Gonina, Hardt, Recht,
  and Talwalkar]{li2020massively}
Li, L.; Jamieson, K.; Rostamizadeh, A.; Gonina, E.; Hardt, M.; Recht, B.;
  Talwalkar, A.
\newblock A System for Massively Parallel Hyperparameter Tuning,  2020.
\newblock arXiv:1810.05934.

\bibitem[Liaw \em{et~al.}(2018)Liaw, Liang, Nishihara, Moritz, Gonzalez, and
  Stoica]{liaw2018tune}
Liaw, R.; Liang, E.; Nishihara, R.; Moritz, P.; Gonzalez, J.E.; Stoica, I.
\newblock Tune: A research platform for distributed model selection and
  training.
\newblock {\em arXiv preprint arXiv:1807.05118} {\bf 2018}.

\bibitem[Kingma and Ba(2014)]{kingma2014adam}
Kingma, D.P.; Ba, J.
\newblock Adam: A method for stochastic optimization.
\newblock {\em arXiv preprint arXiv:1412.6980} {\bf 2014}.

\end{thebibliography}

\end{adjustwidth}

\end{document}